\documentclass[a4paper,fleqn]{cas-dc}

\usepackage[numbers,sort&compress]{natbib}
\usepackage{amsthm}
\usepackage{booktabs}
\usepackage{multirow}
\usepackage{bbding}
\usepackage{pifont}
\newtheorem{proposition}{Proposition}
\usepackage{float}
\usepackage{relsize}
\usepackage{xcolor}

\def\tsc#1{\csdef{#1}{\textsc{\lowercase{#1}}\xspace}}
\tsc{WGM}
\tsc{QE}
\tsc{EP}
\tsc{PMS}
\tsc{BEC}
\tsc{DE}
\begin{document}
\let\WriteBookmarks\relax
\def\floatpagepagefraction{1}
\def\textpagefraction{.001}
\shorttitle{Efficient Redundancy Reduction}
\shortauthors{L Chen et~al.}

\title [mode = title]{Efficient Redundancy Reduction for Open-Vocabulary Semantic Segmentation}                    
\tnotemark[1]

\tnotetext[1]{This research was supported by the National Natural Science Foundation of China under Grant 62306310 and 62433003.}

\author[1,2]{Lin Chen}[
    orcid=0000-0001-6965-1459,
]
\ead{chenlin2024@ia.ac.cn}

\author[1,2]{Qi Yang}[
    orcid=0000-0001-8373-6096,
]
\ead{yangqi2021@ia.ac.cn}

\author[1]{Kun Ding}[
    orcid=0000-0002-2256-8815,
]
\ead{kun.ding@ia.ac.cn}
\cormark[1]

\author[3]{Zhihao Li}
\ead{sdulizh@mail.sdu.edu.cn}

\author[4]{Gang Shen}
\ead{shengang3@chinatowercom.cn}

\author[4]{Fei Li}
\ead{lifei123457@chinatowercom.cn}

\author[1,2]{Qiyuan Cao}
\ead{caoqiyuan2021@ia.ac.cn}

\author[1,2]{Shiming Xiang}[
    orcid=0000-0002-2089-9733,
] 
\ead{smxiang@nlpr.ia.ac.cn}

\affiliation[1]{
    organization={State Key Laboratory of Multimodal Artificial Intelligence Systems (MAIS), Institute of Automation, Chinese Academy of Sciences},
    addressline={No. 95 Zhongguancun East Road},
    city={Beijing},
    postcode={100190},
    country={China}
}

\affiliation[2]{
    organization={School of Artificial Intelligence, University of Chinese Academy of Sciences},
    addressline={No. 19A Yuquan Road},
    city={Beijing},
    postcode={100049},
    country={China}
}

\affiliation[3]{
    organization={School of Software, Shandong University},
    addressline={No. 1500 Shunhua Road},
    city={Jinan},
    postcode={250101},
    country={China}
}

\affiliation[4]{
    organization={China Tower Corporation Limited},
    addressline={No. 9 Dongran North Street},
    city={Beijing},
    postcode={100089},
    country={China}
}

\cortext[cor1]{Corresponding author: Kun Ding (kun.ding@ia.ac.cn)}

\begin{abstract}
Open-vocabulary semantic segmentation (OVSS) is an open-world task that aims to assign each pixel within an image to a specific class defined by arbitrary text descriptions. While large-scale vision-language models have shown remarkable open-vocabulary capabilities, their image-level pretraining limits effectiveness on pixel-wise dense prediction tasks like OVSS. Recent cost-based methods narrow this granularity gap by constructing pixel-text cost maps and refining them via cost aggregation mechanisms. Despite achieving promising performance, these approaches suffer from high computational costs and long inference latency. In this paper, we identify two major sources of redundancy in the cost-based OVSS framework: redundant information introduced during cost maps construction and inefficient sequence modeling in cost aggregation. To address these issues, we propose ERR-Seg, an efficient architecture that incorporates Redundancy-Reduced Hierarchical Cost maps (RRHC) and Redundancy-Reduced Cost Aggregation (RRCA). Specifically, RRHC reduces redundant class channels by customizing a compact class vocabulary for each image and integrates hierarchical cost maps to enrich semantic representation. RRCA alleviates computational burden by performing both spatial-level and class-level sequence reduction before aggregation. Overall, ERR-Seg results in a lightweight structure for OVSS, characterized by substantial memory and computational savings without compromising accuracy. Compared to previous state-of-the-art methods on the ADE20K-847 benchmark, ERR-Seg improves performance by $5.6\%$ while achieving a 3.1× speedup. {The project page is available at \href{https://lchen1019.github.io/ERR-Seg}{https://lchen1019.github.io/ERR-Seg}.}
\end{abstract}

% \begin{graphicalabstract}
% \includegraphics{figs/cas-grabs.pdf}
% \end{graphicalabstract}

% \begin{highlights}
% \item ERR-Seg achieves efficient open-vocabulary semantic segmentation via redundancy reduction.
% \item Quantitative and qualitative analyses confirm the impact of redundancy.
% \item Redundancy-Reduced Hierarchical Cost maps customized lightweight hierarchical cost maps for each image.
% \item Redundancy-Reduced Cost Aggregation reduces computation by $49.3\%$ without performance loss.
% \item ERR-Seg improves performance by $5.6\%$ while achieving a 3.1× speedup on ADE20K-847.
% \end{highlights}


\maketitle

\section{Introduction}
Semantic segmentation is a fundamental task in computer vision that aims to achieve pixel-level semantic understanding. Modern semantic segmentation methods~\cite{chen2017rethinking, xie2021segformer, cheng2022masked, jain2023oneformer, fu2019dual, strudel2021segmenter, chen2018encoder, yuan2020object, chen2022vision, xia2024vit} rely heavily on large-scale labeled datasets, but can only achieve understanding of limited predefined categories. In contrast, humans can understand scenes in an open-vocabulary manner, enabling the comprehension of arbitrary categories. The pursuit of open-vocabulary semantic segmentation (OVSS) is to replicate this intelligence that can perform unbounded pixel-level semantic understanding. Encouragingly, through natural language supervision, recent large-scale visual-language pre-training models (VLMs) exhibit open-vocabulary comprehension ability. However, existing VLMs receive image-level supervision, which is inadequate for the pixel-level representation required by semantic segmentation. This representation granularity gap impedes the transfer of open-world knowledge from VLMs to OVSS.

\begin{figure}[t]
\centering
\includegraphics[width=0.85\columnwidth]{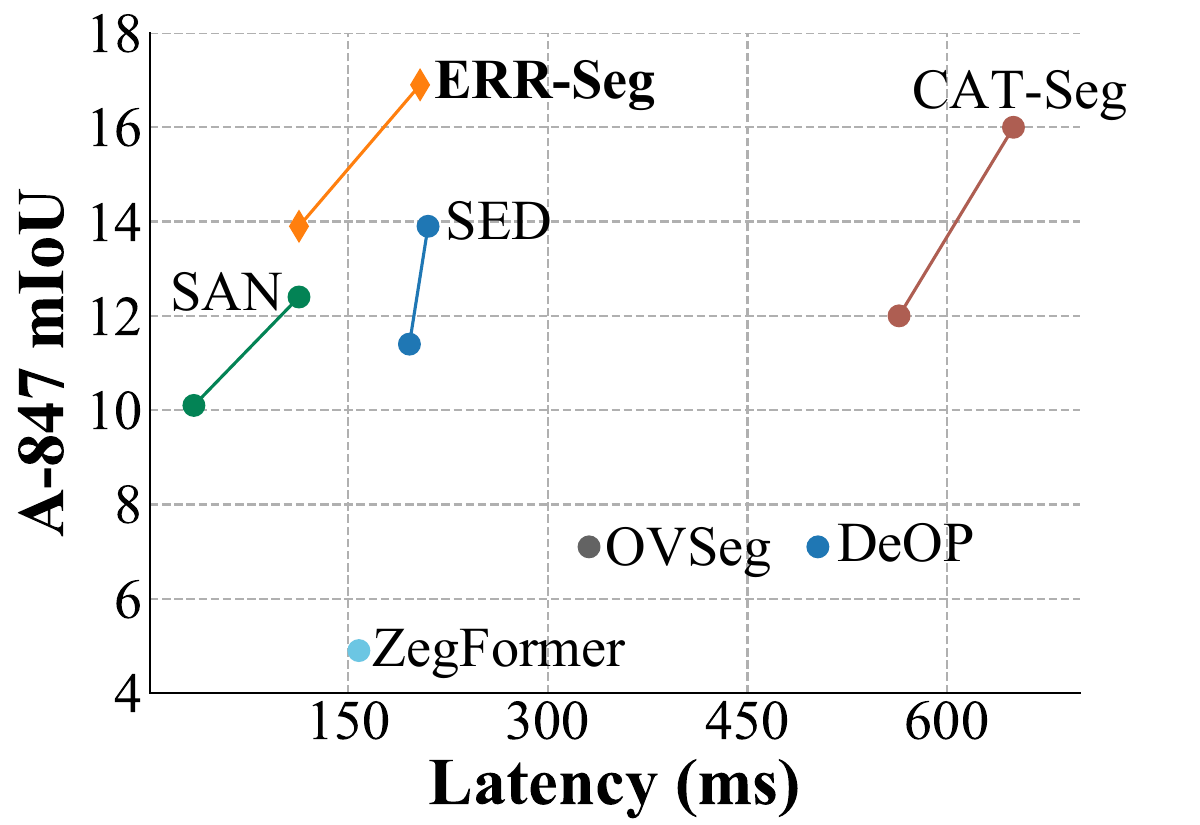}
\caption{Performance vs. latency on ADE20K-847. Compared with ZegFormer~\cite{ding2022decoupling}, OVSeg~\cite{liang2023open}, DeOP~\cite{han2023open}, SAN~\cite{xu2023side}, SED~\cite{xie2024sed} and CAT-Seg~\cite{cho2024cat}, ERR-Seg achieves a new state-of-the-art with lower latency.}
\vspace{-0.4cm}
\label{fig:time_miou}
\end{figure}

For this issue, recent efforts can be categorized into mask-based and cost-based methods. Mask-based methods~\cite{ding2022decoupling,xu2022simple,liang2023open,dong2023maskclip,xu2023open, xu2023side,yu2024convolutions, shan2024open, zhu2024mrovseg, chen2024sam} conceptualize OVSS as the combination of mask generation and open-vocabulary mask classification. These methods necessitate region-level representation, which is much closer to the image-level representation. However, mask-based methods depend on a region proposal network (RPN) to generate masks, which is typically trained on a specific dataset, lacking the region proposal capability in the open-vocabulary context. To address this issue, several studies~\cite{chen2023semantic, ren2024grounded, chen2024sam, wang2024sam} employ SAM~\cite{kirillov2023segment} as a universal RPN. Nevertheless, as SAM tends to over-segment an instance into sub-patches, these methods have not made significant progress yet. In contrast, cost-based methods~\cite{cho2024cat,xie2024sed} directly generate coarse-grained cost maps by calculating the cosine similarity between dense pixel-level embeddings and text embeddings. These methods effectively bridge the representation granularity gap and eliminate the requirement of RPN. As a result, the methods in this family have shown superior performance compared to mask-based methods.

Although cost-based methods have effectively addressed the challenge of representation granularity, they introduce cost aggregation over pixel-text cost maps, incurring significant computational overhead. Moreover, this overhead grows quadratically with the number of categories, in contrast to the linear growth of mask-based methods. We attribute this issue to two inherent design redundancies in the cost-based methods: (1) \textbf{Redundancy in Cost Maps}: It is evident that real-world categories are limitless, while those depicted in an image are always limited, leading to numerous redundant classes. As illustrated in Table~\ref{tab:avg}, images annotated with $847$ classes have an average of only $10.2$ categories, indicating that there are $98.8\%$ redundant classes. On the one hand, it leads to increased latency due to the redundant class dimension of the cost maps. On the other hand, it poses a challenge for the attention mechanism to model class-level sequences containing numerous redundant classes. (2) \textbf{Redundancy in Cost Aggregation}: In closed-set semantic segmentation~\cite{fu2019dual, yuan2020object, xie2021segformer}, the fixed class vocabulary enables attention mechanisms to jointly model spatial-class and class-level contextual information with class-informative channels. In contrast, the cost-based OVSS methods require class-agnostic channels due to dynamic class vocabulary, leading to the spatial-level and class-level decoupled attention mechanism in cost aggregation. While effective, it incurs significant computational and memory overhead. Additionally, due to the decoupling, the attention mechanism's objective is relatively straightforward, potentially containing redundant computations that can be reduced.

\begin{figure}[t]
\centering
\includegraphics[width=0.97\columnwidth]{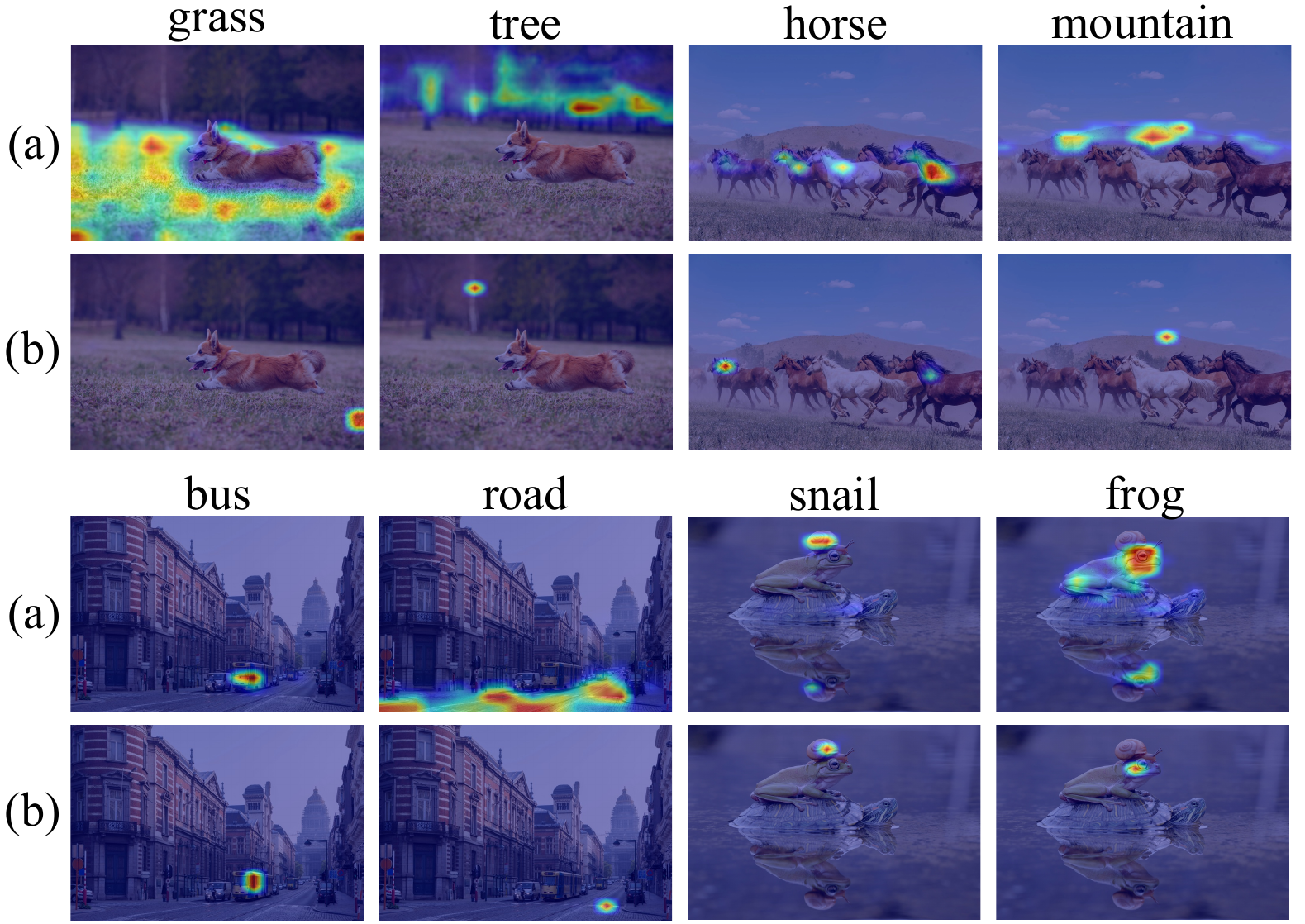}
\vspace{-3pt}
\caption{Visual comparison of attention maps during cost aggregation. (a) Using our proposed cost maps (RRHC) with redundancy reduction (48 classes) and (b) using the original full cost maps (847 classes). The results show that the attention mechanism more effectively captures spatial-level long-range dependencies in our proposed RRHC.}
\vspace{-15pt}
\label{fig:RCR_visual}
\end{figure}

To this end, we propose an efficient framework in view of redundancy reduction for OVSS, jointly considering the efficiency and accuracy. To alleviate the redundancy (1), ERR-Seg presents the Redundancy-Reduced Hierarchical Cost maps (RRHC). It first selects the most relevant categories while discarding the others based on the category existence probability implied in the pixel-text cost maps. Additionally, it transforms middle-layer features from CLIP's image encoder into semantic space to compute hierarchical cost maps in parallel, which are then merged to aggregate more middle-layer semantic details. By reducing the channel dimensions, the attention mechanism can better capture spatial-level long-range dependencies (as depicted in Fig.~\ref{fig:RCR_visual}), while also significantly decreasing latency. More comprehensive analyses are provided in Sec.~\ref {sec:analysis}. To alleviate the redundancy (2), ERR-Seg introduces Redundancy-Reduced Cost Aggregation (RRCA), which incorporates spatial-level and class-level sequence reduction before cost aggregation. This results in a $49.3\%$ reduction in computational workload in cost aggregation without compromising performance. As demonstrated in Fig.~\ref{fig:time_miou}, ERR-Seg establishes a new state-of-the-art in terms of efficiency and accuracy.

The main contributions can be summarized as follows:
\begin{itemize}
    \item {We introduce an efficient cost-based OVSS framework in view of redundancy reduction, which achieves superior performance on multiple datasets and significantly accelerates the inference process.}
    \item We introduce Redundancy-Reduced Hierarchical Cost maps (RRHC), characterized by dynamic redundant class reduction for each image and hierarchical cost maps extracted from middle-layer features.
    \item We present the Redundancy-Reduced Cost Aggregation (RRCA) with spatial-level and class-level sequence reduction, substantially accelerating the aggregation process without compromising accuracy.
    \item We quantitatively and qualitatively investigate the impact of redundancy reduction on modeling spatial-level and class-level contextual information, as well as its effect on computational complexity.
\end{itemize}

\begin{table}[t]
\centering
\small
\caption{Analysis of image category count across five semantic segmentation datasets. It reveals that both PAS-20 (with 20 categories) and A-847 (with 847 categories) contain more than $90\%$ redundant classes.}
\resizebox{\columnwidth}{!}{
\begin{tabular}{l|ccccc}
\toprule
\textbf{} & PAS-20 & PC-59 & A-150 & PC-459 & A-847 \\ \midrule
Average      & 1.5         & 4.8   & 8.5   & 6.5    & 10.2  \\ 
Redundancy   & $92.5\%$    & $91.9\%$   & $94.3\%$   & $98.6\%$    & $98.8\%$  \\ \bottomrule
\end{tabular}
}
\label{tab:avg}
\end{table}

\section{Related Work}
\subsection{Closed-set Semantic Segmentation}
{From the perspective of modeling semantic segmentation tasks, technical methods can be divided into two categories: (1) Pixel-based methods consider the semantic segmentation task as a dense pixel-by-pixel prediction task. FCN~\cite{long2015fully} serves as a fundamental pixel-based approach by substituting the fully connected layer in conventional CNNs with a convolutional layer to enable dense category predictions. Subsequently, various pixel-based variants have been developed to incorporate more contextual information, such as dilated convolution~\cite{chen2017rethinking, chen2018encoder}, feature pyramid~\cite{kirillov2019panoptic, xiao2018unified}, pyramid pooling module~\cite{zhao2017pyramid, xiao2018unified}, object contextual representation~\cite{yuan2020object}, attention mechanisms~\cite{fu2019dual, huang2019ccnet}, pixel contrast~\cite{wang2021exploring}, prototype view~\cite{zhou2022rethinking} and hierarchical semantic~\cite{li2022deep, li2023logicseg}. Additionally, some methods~\cite{strudel2021segmenter, zheng2021rethinking, xie2021segformer, gu2022multi, chen2022vision, xia2024vit, liang2023clustseg} introduce transformer-based architectures to semantic segmentation. (2) Mask-based methods~\cite{cheng2021per, zhang2021k, cheng2022masked, jain2023oneformer} treat the segmentation task as a combination of mask generation and mask classification. These methods aim to unify various segmentation tasks (such as instance segmentation, semantic segmentation, and panoptic segmentation) and have demonstrated their effectiveness in the context of closed-set semantic segmentation.}

As a fundamental task in computer vision, closed-set semantic segmentation has achieved remarkable success in the past decade, particularly in specific domains such as remote sensing~\cite{ma2024sam, geng2023dual}, medical segmentation~\cite{zhou2024uncertainty} and autonomous driving scenarios~\cite{wu2024s, chang2023multi}. However, this approach relies heavily on large-scale labeled datasets, which require significant human effort to create. Moreover, it is limited to recognizing only a predefined set of categories. This study focuses on OVSS, which can understand arbitrary vocabulary lists, allowing it to address more complex and dynamic scenarios.

\begin{figure*}[t]
\centering
\includegraphics[width=2\columnwidth]{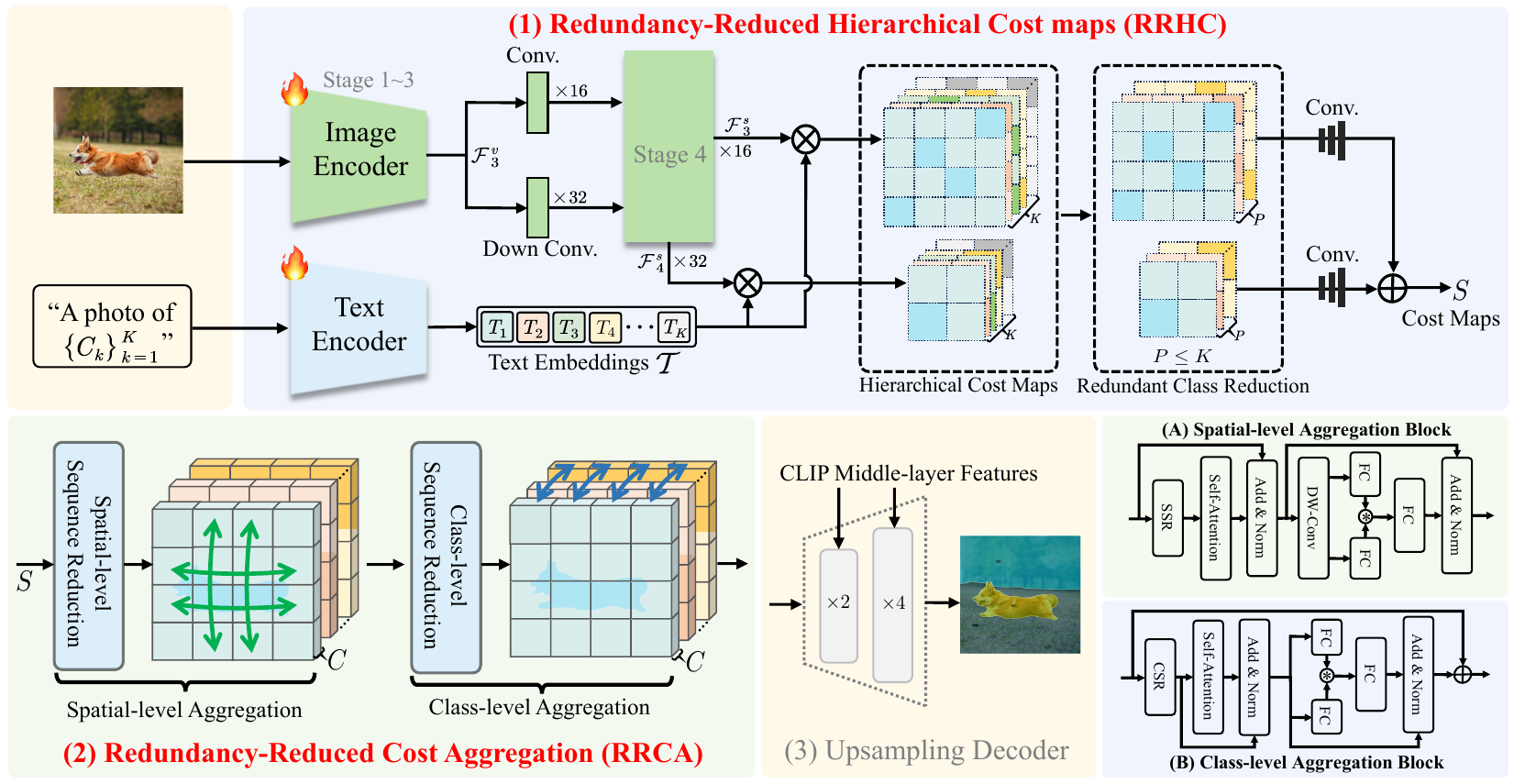}
\caption{Overall architecture of ERR-Seg. Initially, redundancy-reduced hierarchical cost maps are generated by extracting cost maps from middle-layer features and eliminating class redundancy. Subsequently, the sequence length is reduced before cost aggregation to speed up the computation. Finally, the upsampling decoder restores the high-rank information of cost maps by incorporating image details from the middle-layer features of CLIP's visual encoder.}
\label{fig:overview}
\end{figure*}

\subsection{Vision-Language Models}
Visual-language models (VLMs) aim to extend open-world comprehension to the visual domain by jointly training visual and language modalities. Early VLMs~\cite{chen2019uniter, su2019vl, lu2019vilbert, li2020oscar} were trained on small-scale datasets through multiple multi-modal tasks. In contrast, CLIP~\cite{radford2021learning} collects large-scale image-text pairs and uses image-text contrastive learning to align the visual and language modality. To further enhance the quality and efficiency of image-text pre-training, ALIGN~\cite{jia2021scaling} collected a much larger dataset of image-text pairs without removing noisy data, SigLIP~\cite{zhai2023sigmoid} introduces pairwise sigmoid loss to eliminate the requirement of a global view of the pairwise similarities for normalization. Beyond image-text contrast learning, BLIP-2~\cite{li2023blip} and LLaVA~\cite{liu2024visual} introduce an additional language model to process aligned images and text in a unified framework for more complex reasoning. 

VLMs have been successfully utilized in various multi-modal downstream tasks such as text-to-image generation~\cite{ramesh2022hierarchical, zhang2023adding, wang2024prior}, referring segmentation~\cite{wang2022cris, shang2022cross}, cross-modal retrieval~\cite{cheng2022semantic, li2024ckdh} and open-vocabulary understanding~\cite{liu2024grounding, yu2024convolutions, xie2024sed}. In this study, we better support the application of VLMs to OVSS by enhancing their ability to generate hierarchical semantic maps and leveraging the prior knowledge from VLMs for redundancy reduction.

\subsection{Open-Vocabulary Semantic Segmentation}
Earlier methods~\cite{bucher2019zero, xian2019semantic, zhao2017open} for OVSS attempt to establish a joint embedding space connecting image pixels and a predefined vocabulary list. However, these methods struggle with arbitrary class recognition due to their limited vocabulary size. Fortunately, vision-language models (VLM)~\cite{radford2021learning, jia2021scaling, li2022blip} pre-trained on extensive web data exhibit strong capabilities in open-vocabulary recognition,  leading recent research focused on adapting these models for OVSS. Based on the method of introducing semantic information from VLM, current OVSS methods can be categorized into two main groups: mask-based methods and cost-based methods.

% Mask-based method
For mask-based methods, an intuitive two-stage approach is first proposed. Initially, an additional RPN is adopted to propose region candidates, followed by the application of VLMs for open-vocabulary recognition. ZegFormer~\cite{ding2022decoupling} and SimBaseline~\cite{xu2022simple} directly adopt this two-stage method, leading to suboptimal performance and significant computational overload. To improve the performance, ODISE~\cite{xu2023open} utilizes a text-to-image diffusion model as RPN to generate high-quality masks, and OVSeg~\cite{liang2023open} fine-tunes CLIP on region-text pairs for improved open-vocabulary recognition. To further reduce the computational overload, single-stage methods have been introduced to avoid multiple forward propagations. Among these methods, MaskCLIP~\cite{ding2022open} and SAN~\cite{xu2023side} adopt attention bias, while FC-CLIP~\cite{yu2024convolutions} employs mask pooling. {Recent works have introduced various techniques to enhance performance. EBSeg~\cite{shan2024open} alleviates overfitting to seen training classes via image embedding balancing. MROVSeg~\cite{zhu2024mrovseg} tackles the resolution discrepancy by introducing a multi-resolution training framework. SCAN~\cite{liu2024scan} and MAFT+~\cite{jiao2024maftp} enhance the fusion of visual and textual features through semantic guidance and a collaborative optimization mechanism, respectively. More recently, OVSNet~\cite{liu2025stepping} further avoids overfitting by fusing heterogeneous features and expanding the training space.}

% cost-based method
For cost-based methods, CAT-Seg~\cite{cho2024cat} aggregates the pixel-text cost maps at both the spatial-level and class-level at one resolution. It also emphasizes the importance of fine-tuning the text encoder of CLIP. In contrast, SED~\cite{xie2024sed} gradually aggregates the cost maps at different resolutions. However, cost-based methods have a drawback compared to mask-based methods: they involve operators with quadratic complexity in relation to the vocabulary size. This means that when the vocabulary is large, it can lead to significant computational and storage overhead. Our proposed ERR-Seg is also cost-based, but we propose to reduce redundancy to alleviate this issue.

\section{Methodology}
Figure \ref{fig:overview} presents an overview of our proposed ERR-Seg. To address the inherent redundancy within the cost-based framework for OVSS, we propose eliminating redundancies in both cost maps and cost aggregation. This section details our redundancy reduction approach for cost maps in Sec.~\ref{sec:rrh}, followed by the method for reducing redundancy in cost aggregation in Sec.~\ref{sec:agg}. Additionally, Sec.~\ref{sec:decoder} provides a brief introduction to the upsampling encoder adopted in the cost-based framework. Finally, Sec.~\ref{sec:analysis} presents a qualitative and quantitative analysis of the impact of redundancy.

\subsection{Redundancy-Reduced Hierarchical Cost Maps}
\label{sec:rrh}
In this section, we present the methodology for constructing our proposed Redundancy-Reduced Hierarchical Cost maps (RRHC). It diverges from previous cost maps in two aspects: (1) Hierarchical Cost Maps: previous methods typically utilize only the last layer output from the CLIP's image encoder, whereas RRHC also incorporates the features from intermediate layers to build hierarchical cost maps. (2) Redundant Class Reduction: we eliminate redundant classes to derive more concise cost maps customized for each image.

\subsubsection{Hierarchical Cost Maps}
Typically, cost-based methods introduce semantic information by calculating pixel-text cost maps using only the last layer output, which ignores semantic details from middle-layer features. To address this, we extract semantic details from the middle-layer features from the CLIP's image encoders to build hierarchical cost maps.

Initially, $I$ is first encode to derive the middle-layer features $\mathcal{F}_i^v, i\in \{1,2,3\}$, and dense image-text aligned embeddings $\mathcal{F}_4^s$. The relationship between $\mathcal{F}_3^v$ and $\mathcal{F}_4^s$ is modeled as follows:
\begin{align}
  \mathcal{F}_4^s&=f_{\mathrm{head}}\left( g\left( conv_{s=2}(\mathcal{F}_3^v) \right) \right),
\end{align}
where $g$ represents stage 4 of CLIP's visual encoder, $f_{\mathrm{head}}$ is the linear projection head, and $conv_{s=2}$ denotes convolution with a stride of 2 for downsampling. Additionally, we generate extra image-text aligned embeddings from middle-layer features $\mathcal{F}_3^v$, as follows:
\begin{align}
  \mathcal{F}_3^s=f_{\mathrm{head}}\left( g \left( conv_{s=1}(\mathcal{F}_3^v) \right) \right),
\end{align}
where $conv_{s=1}$ denotes convolution with a stride of 1, using the same initial values as $conv_{s=2}$. $conv_{s=1}$ prevents downsampling, thereby keeping the size of $\mathcal{F}_3^s$ the same as $\mathcal{F}_3^v$. Notably, we do not construct cost maps from $\mathcal{F}_1^v$ or $\mathcal{F}_2^v$, as these high-resolution features primarily capture high-frequency details. Therefore, incorporating them during the upsampling stage is sufficient.

Subsequently, by giving $\mathcal{F}_3^v \in \mathbb{R}^{C \times \frac{H}{16} \times \frac{W}{16}}$ and $\mathcal{F}_4^v \in \mathbb{R}^{C \times \frac{H}{32} \times \frac{W}{32}}$, we computes the hierarchical cost maps. Given a vocabulary list $\mathcal{C}=\{C_1, ..., C_K\}$, a set of sentences $\mathcal{G}=\{ G_1,...,G_K \}$ is first generated, where $G_i$ consists of $C_i$ and a well-crafted prompt (e.g., ``a photo of $\{C_i\}$ in the scene"). $\mathcal{G}$ is then fed into the CLIP's text encoder to get text embeddings for each class, denoted as $\mathcal{T}=\{ T_1,..., T_K \}$. The pixel-text cost maps are derived by calculating the cosine similarity between each pixel and each class:
\begin{align}
    S(i,j,t)=\frac{\mathcal{F}^s(i,j)\cdot \mathcal{T}(t)}{\left\| \mathcal{F}^s(i,j) \right\| \left\| \mathcal{T}\left( t \right) \right\|},
\end{align}
where $(i, j)$ denotes the 2D spatial positions, and $t$ represents the index of text embeddings, $\mathcal{F}^s = \mathcal{F}_3^s $ or $\mathcal{F}_4^s$. The $\mathcal{F}_3^s$ and $\mathcal{F}_4^s$ are used to derive the pixel-text cost maps $S_3 \in \mathbb{R}^{K \times \frac{H}{16} \times \frac{W}{16}}$ and $S_4 \in \mathbb{R} ^ {K \times \frac{H}{32} \times \frac{W}{32}}$ individually, combining into our proposed hierarchical cost maps.

\begin{figure}[t]
\centering
\includegraphics[width=\columnwidth]{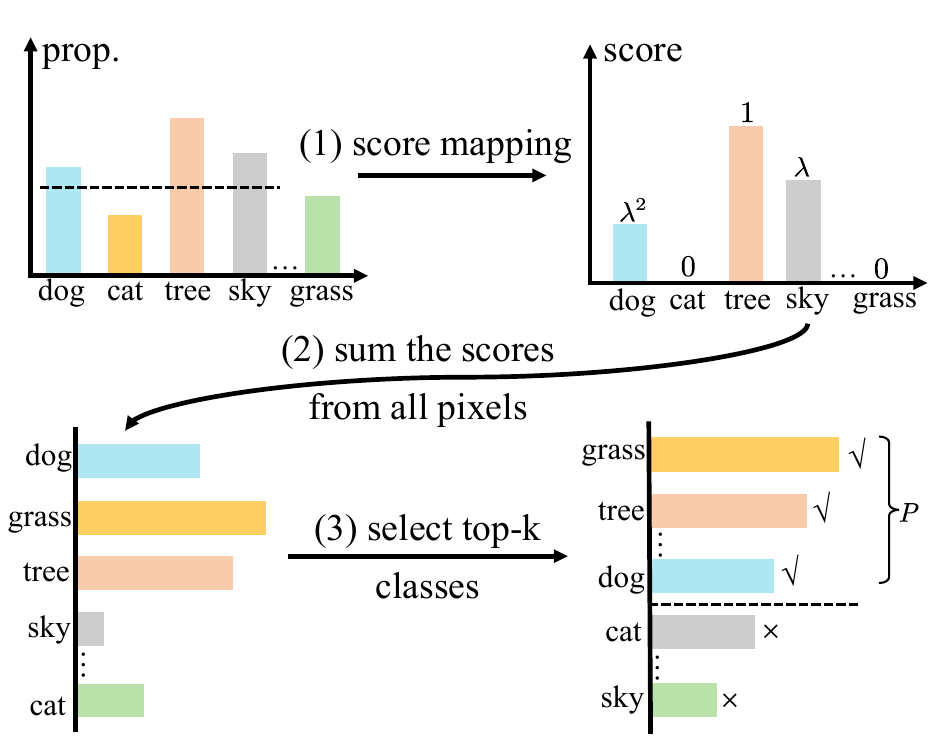}
\caption{Pipeline of our proposed redundant class reduction mechanism. It involves a training-free scoring function to assign scores to each class, retaining the top-$P$ classes while eliminating other redundant classes.
}
\label{fig:RCR_pipline}
\end{figure}

\subsubsection{Redundant Class Reduction}
Based on the observation that the number of classes within an image is limited, we introduce a training-free class selection function $\Phi$, which leverages the prior knowledge in the original cost maps to reduce redundant classes, and determines the top-$P$ classes. Here, $P \le K$ and $P$ is an empirical value. The impact of $P$ on training and inference will be discussed later.

The pipeline of the function $\Phi$ is illustrated in Fig.~\ref{fig:RCR_pipline}. $S$ represents the probability of each category at each spatial location. Subsequently, for each spatial location, the top $k$ classes are selected based on their probability values and arranged in descending order:
\begin{align}
    N = Top(S, k),
\end{align}
where $N \in \mathbb{R}^{k \times H_c \times W_c}$ represents the indices of the top $k$ categories, $H_c \times W_c$ is the size of input cost maps. Then we calculate the scores for each category. The score for $t$-th category, denoted as $R(t)$, is computed as:
\begin{align}
    R(t)=\sum_{c=1}^{k}{\lambda ^ {c-1} \left( \sum_{i=1}^{H_c}{\sum_{j=1}^{W_c}{\mathbb{I} \left[N(c, i, j)=t \right]}} \right)},
\end{align}
where $\lambda ^ {c-1}$ represents the score assigned to the class ranked $c$. $\lambda$ is the score coefficient and is expected to meet the condition $0< \lambda <1$, as higher-ranked categories should receive higher scores. The notation $\mathbb{I}[\cdot]$ represents a logical expression that yields 1 when the condition is true and 0 otherwise. The cost map used in $\Phi$ is $S_4$, since it is built from the origin final output of CLIP's image encoder. After scoring all categories, we obtain $\left\{ R(i) \right\} _{i=1}^{K}$, only the top $P$ categories are retained, thus forming redundancy-reduced $S_3$ and $S_4$.

\subsubsection{Cost Maps Embedding and Merging}
After removing redundant class, we employ a convolution layer with input channel 1 and output channel $C$ to embed the cost maps $S_3$ and $S_4$ into the latent space. To aggregate the hierarchical cost maps, we employ a feature pyramid strategy. Specifically, it can be computed as:
\begin{align}
S \leftarrow \textrm{conv}\left( S_3 \right) + \textrm{up} \left( \textrm{conv}\left( S_4 \right) \right),
\end{align}
where $\mathrm{conv}$ denotes a convolution layer, $\mathrm{up}$ represents upsampling interpolation to ensure the same size. Subsequently add them to yield final cost maps $S \in \mathbb{R}^{P \times C \times H_s \times W_s}$, where $H_s = \frac{H}{16}$ and $W_s = \frac{W}{16}$.

\subsection{Redundancy-Reduced Cost Aggregation}
\label{sec:agg}
The cost-based OVSS framework employs spatial-level and class-level decoupled aggregation to enhance contextual information in the coarse-grained cost maps $S$. Although this aggregation is necessary and effective, it incurs significant computational redundancy due to the attention mechanism modeled across extensive sequence dimensions. Therefore, we introduce Spatial-level Sequence Reduction (SSR) and Class-level Sequence Reduction (CSR) strategies before the aggregation stages. SSR and CSR selectively compress the sequence lengths involved in attention computation, significantly accelerating context fusion while maintaining the segmentation accuracy.

\subsubsection{Spatial-level Aggregation}
Spatial-level aggregation aims to model the relationships between different spatial locations within the same class. The architecture of the spatial-level aggregation block is presented in Fig.~\ref{fig:overview} (a). The cost maps $S$ is first reshaped into a spatial-level sequence $S_{\mathrm{spa}} \in \mathbb{R}^{P \times (H_s \times W_s) \times C}$ and then adopt self-attention mechanism, where the sequence length is $H_s \times W_s$. Notably, since the class vocabulary is dynamic in OVSS, the channel dimension $C$ is inherently class-agnostic, thus self-attention operates independently on each class sequence. Although this approach can effectively model intra-class relations, it requires computing self-attention over all $H_s \times W_s$ pixels for every class, leading to substantial computational redundancy.

Recognizing the inherent redundancy in modeling dense interactions across all $H_s \times W_s$ pixels per class, we introduce Spatial-level Sequence Reduction (SSR) to reduce the computational complexity in spatial-level aggregation. Specifically, it reduces the sequence length of $S$ before reshaping into a spatial-level sequence, as follows:
\begin{align}
    S^{r} = f_{\mathrm{proj}} \left( \mathrm{conv}_{s=r_1}(S) \right),
\end{align}
where $S^{r} \in \mathbb{R}^{P \times (\frac{H_s}{r_1} \times \frac{W_s}{r_1}) \times C}$, $\mathrm{conv}_{s=r_1}$ is a convolution layer with stride $r_1$ that performs downsampling on the cost maps while increasing the channel dimension, and $f_{\mathrm{proj}}$ represents a linear projection layer that restores the original channel dimension. Subsequently, in the self-attention module, the original sequence $S_{\mathrm{spa}}$ serves as query to preserve output dimensions, and the compressed sequence $S_{\mathrm{spa}}^{r}$ functions as key and value.

The output of the self-attention module $S_{\mathrm{spa}}$ then passes through a depth-wise convolution layer, followed by a two-layer MLP. Considering SSR compresses the spatial dimension for computational efficiency, it is necessary to enhance the feature representation capacity within the MLP. Therefore, we replace the first linear layer of the MLP with a star operation~\cite{ma2024rewrite}, leveraging its ability to effectively map features into a higher-dimensional space for richer interactions. Specifically, the modified MLP layer is defined as:
\begin{align}
    S_{\mathrm{spa}} \leftarrow f_{\mathrm{out}}\left( \sigma \left( f_{\mathrm{in}}^1(S_{\mathrm{spa}}) \odot f_{\mathrm{in}}^2 (S_{\mathrm{spa}}) \right) \right),
\end{align}
where, $\sigma$ denotes the activation function, $f_{\mathrm{in}}^1$ and $f_{\mathrm{in}}^2$ are linear projections mapping from $\mathbb{R}^C$ to $\mathbb{R}^{C'}$, $\odot$ represents element-wise multiplication, $f_{\mathrm{out}}$ is the output projection mapping back to $\mathbb{R}^C$, and $C'$ is the expanded intermediate dimension.

\subsubsection{Class-level Aggregation}
Class-level aggregation aims to capture dependencies among different classes within the same spatial location. The architecture of our proposed class-level aggregation block is depicted in Fig.~\ref{fig:overview} (b). It begins by reshaping the cost maps $S$ into a class-level sequence $S_{\mathrm{cls}} \in \mathbb{R}^{(H_s \times W_s) \times P \times C}$, where $P$ is the sequence length for the subsequent self-attention operation. The goal of this self-attention mechanism is to model inter-class relationships, facilitating the identification of the most relevant category for each pixel based on cross-class interactions. However, performing this attention computation over all $H_s \times W_s$ pixels per spatial location is computationally intensive and potentially redundant.

Recognizing that neighboring pixels often share similar features and thus exhibit correlated class dependencies, we introduce the Class-level Sequence Reduction (CSR) strategy. CSR applies an average pooling operation with kernel size $r_2 \times r_2$ and stride $r_2$ to spatially downsample the cost maps before reshaping. This reduces the spatial dimensions to $\frac{H_s}{r_2} \times \frac{W_s}{r_2}$. Consequently, the reshaped class-level sequence becomes $S_{\mathrm{cls}} \in \mathbb{R}^{(\frac{H_s}{r_2} \times \frac{W_s}{r_2}) \times P \times C}$. Subsequently, the modified MLP in spatial-level aggregation is also employed for improving feature representation capabilities. Finally, the output features are upsampled to the original spatial resolution ($H_s \times W_s$) and added element-wise to the input cost map $S$. This compression offers dual benefits: (1) it aggregates information from adjacent pixels into a single representation, providing richer context for category identification at class-level aggregation, and (2) it drastically reduces the sequence length involved in the self-attention computation from $H_sW_s$ to $\frac{H_sW_s}{r_2^2}$, significantly improving computational efficiency.

\subsection{Upsampling Decoder}
\label{sec:decoder}
After the cost aggregation, the cost maps $S$ incorporate more contextual information but remain at low resolution, lacking fine-grained edge details. To recover high-resolution semantic maps, we leverage middle-layer visual features $\mathcal{F}_i^v$ extracted from the CLIP image encoder to guide the upsampling of $S$, following existing practices~\cite{cho2024cat}. Initially, $\mathcal{F}_i^v$ undergoes a convolution layer and is then concatenated with the cost maps at each class. The concatenated features are then passed through a refinement block, comprising two convolution layers. Finally, high-resolution cost maps $S_{hres} \in \mathbb{R}^{P \times C_h \times \frac{H}{4} \times \frac{W}{4}}$ is obtained, where $C_h$ is the channel dimension of the cost maps, which gradually decreases in the upsampling decoder. Finally, a linear project head project $S_{hres}$ into the size of $P \times (1) \times \frac{H}{4} \times \frac{W}{4}$, to get the final logits.

\subsection{Analysis on Redundancy Reduction}
\label{sec:analysis}
A comprehensive analysis of our proposed redundancy reduction is provided in this section, encompassing three aspects: the quantitative analysis of class-level aggregation, the qualitative features of spatial-level aggregation, and the computational complexity.

\subsubsection{Impact on class-level aggregation}
The original cost maps $S$ comprise the existing and redundant classes. Let $\mathcal{Q}_e=\{S(i)\}_{i=1}^{M}$ collect the original cost maps corresponding to the existing classes, where $M$ is the number of existing classes in $I$. The cost maps corresponding to the redundant classes are represented by $\mathcal{Q}_r=\{S(i)\}_{i=M+1}^{K}$, and those corresponding to the redundant classes after partial reduction are denoted as $\mathcal{Q}_p=\{S(i)\}_{i=M+1}^{P}$. In the subsequent subsections, we will analyze the impact of $\mathcal{Q}_r$ and $\mathcal{Q}_p$ on class-level and spatial-level aggregation (the primary operations on the cost maps), as well as on latency.

The class-level attention output of the $q$-th class in $\mathcal{Q}_e$ before reduction is:
\begin{align}
\scalebox{0.85}{$\displaystyle
O^{e,r}(q)=\sum_{k=1}^K{\exp \left( \frac{A_{qk}}{\sqrt{C}} \right) S(k)}/\sum_{k=1}^K{\exp \left( \frac{A_{qk}}{\sqrt{C}} \right)},
$}
\label{er}
\end{align}
where $q \le M$, and $A \in \mathbb{R}^{K\times K}$ stands for the class-level attention score matrix before reduction. The contribution of $\mathcal{Q}_r$ to the attention output of the $q$-th class is formulated as:
\begin{align}
\scalebox{0.85}{$\displaystyle
O^{r}(q)=\sum_{k=M+1}^K{\exp \left( \frac{A_{qk}}{\sqrt{C}} \right) S(k)}/\sum_{k=1}^K{\exp \left( \frac{A_{qk}}{\sqrt{C}} \right)}.
$}
\label{q}
\end{align}
Similarly, we denote the output of the $q$-th class after reduction as $O^{e,p}(q)$, and the contribution of $\mathcal{Q}_p$ as $O^{p}(q)$.
\begin{proposition}
The relationship between the contribution ratio of $\mathcal{Q}_r$, denoted as $\Delta _r$, and the contribution ratio of $\mathcal{Q}_p$, denoted as $\Delta _p$, is expressed as:
\begin{align}
    \Delta _p=\frac{O^{p}(q)}{O^{e,p}(q)} < \Delta _r=\frac{O^{r}(q)}{O^{e,r}(q)}.
\end{align}
\end{proposition}
The above inequality suggests that as the number of redundant classes increases, they contribute more to the final output. Thus, existing classes contribute less. As such, we can deduce that reducing redundancy facilitates the class-level aggregation to focus more on existing classes. The proof is detailed in the Appendix.

\subsubsection{Impact on spatial-level aggregation}
Spatial-level aggregation aims to capture dependencies among different pixels within the same class, it typically assigns high scores when two pixels are close or distant but belong to the same class, indicating similar embeddings. However, after class-level aggregation, each pixel-level embedding integrates contributions from redundant classes. This leads to decreased similarity among pixel-level embeddings of the same class, hindering spatial-level attention from effectively capturing dependencies on distant pixels within the same class. In Fig.~\ref{fig:RCR_visual}, we visualize the spatial-level attention maps of different classes, confirming our conclusion that redundancy causes spatial-level attention to focus more on local regions and compromises the ability to capture long-range dependencies.

\subsubsection{Impact on computational complexity}
The computational efficiency of ERR-Seg is achieved through two redundancy reduction strategies: redundant class dimension reduction from $K$ to $P$, combined with spatial-level and class-level sequence reduction through downsampling ratios $r_1, r_2$. For spatial-level attention, dimensionality compression transforms the spatial sequence from $K \times H_sW_s \times C$ to $P \times \frac{H_sW_s}{r_1^2} \times C$, reducing complexity from $O(KCH_s^2W_s^2)$ to $O(PCH_s^2W_s^2 / r_1^2)$. For class-level attention, dimensionality compression transforms complexity from $O(K^2CH_sW_s)$ to $O(P^2C \cdot H_sW_s / r_2^2)$. This yields computational savings of $1 - P/(r_1^2K)$ for spatial-level aggregation and $1 - (P/(r_2K))^2$ for class-level aggregation.

\section{Experiments}
\subsection{Experimental Setup}
\subsubsection{Traning dataset}
Following the existing OVSS methods~\cite{cho2024cat, xie2024sed}, we train our model on COCO-Stuff~\cite{caesar2018coco}. It has 118,287 images for training with 171 different classes.

\subsubsection{Evaluation dataset}
\begin{itemize}
    \item \textbf{PASCAL-Context}~\cite{mottaghi2014role} contains 459 classes with 5,104 images for validation. We report the scores of the full classes (PC-459) and the scores of the 59 most frequent classes (PC-59).
    \item \textbf{ADE20K}~\cite{zhou2019semantic} has 2,000 validation images with 847 classes. Similarly to PASCAL-Context, we report the results of 150 frequent classes (A-150) and full 847 classes (A-847).
    \item \textbf{PASCAL VOC}~\cite{everingham2010pascal} includes 1,449 images for validation with 20 common object classes. We report the performance of the 20 classes named PAS-20.
\end{itemize}

\subsubsection{Analysis of the evaluation dataset}
\label{analysis_data}
Prior studies have typically overlooked the label-level similarity between the validation and training sets, while it is necessary for open vocabulary-related tasks. In this section, we extract text embeddings for each category and then calculate the cosine similarity between the categories in the evaluation and training datasets, resulting in a similarity matrix. Subsequently, we map each category in the evaluation dataset to a category with the highest similarity score in the training dataset. The results are presented in Table~\ref{tab:dataset}; we report the proportion of labels in the validation sets with a similarity score equal to $1.00$, greater than $0.95$, $0.90$, and $0.85$.

Most categories in PAS-20 and PC-59 display similarity scores exceeding $0.90$, indicating that a significant number of categories in these datasets were encountered during the training process. This also implies that overfitting to COCO-Stuff can lead to performance improvement in PAS-20 and PC-59. Hence, PAS-20 and PC-59 measure more in-domain open-vocabulary semantic segmentation capability. In contrast, the ratio is comparatively low for A-150, PC-459, and A-847, suggesting that these three datasets are suitable for measuring out-domain open-vocabulary capability.

\begin{table}[H]
\centering
\small
\caption{Label-level similarity between evaluation and training datasets.}
\begin{tabular}{l|cccc}
\toprule
Dataset                  & $1.00$     &  $\ge0.95$     & $\ge0.90$   \\ \midrule
PAS-20                 & 0.45          & 0.90            & 1.00        \\
PC-59                  & 0.30          & 0.85            & 0.95         \\
A-150                & 0.19          & 0.41            & 0.57         \\ 
PC-459                 & 0.10          & 0.21            & 0.29         \\
A-847                & 0.06          & 0.12            & 0.21         \\ \bottomrule
\end{tabular}
\label{tab:dataset}
\end{table}

\begin{figure*}[H]
\centering
\includegraphics[width=1.91\columnwidth]{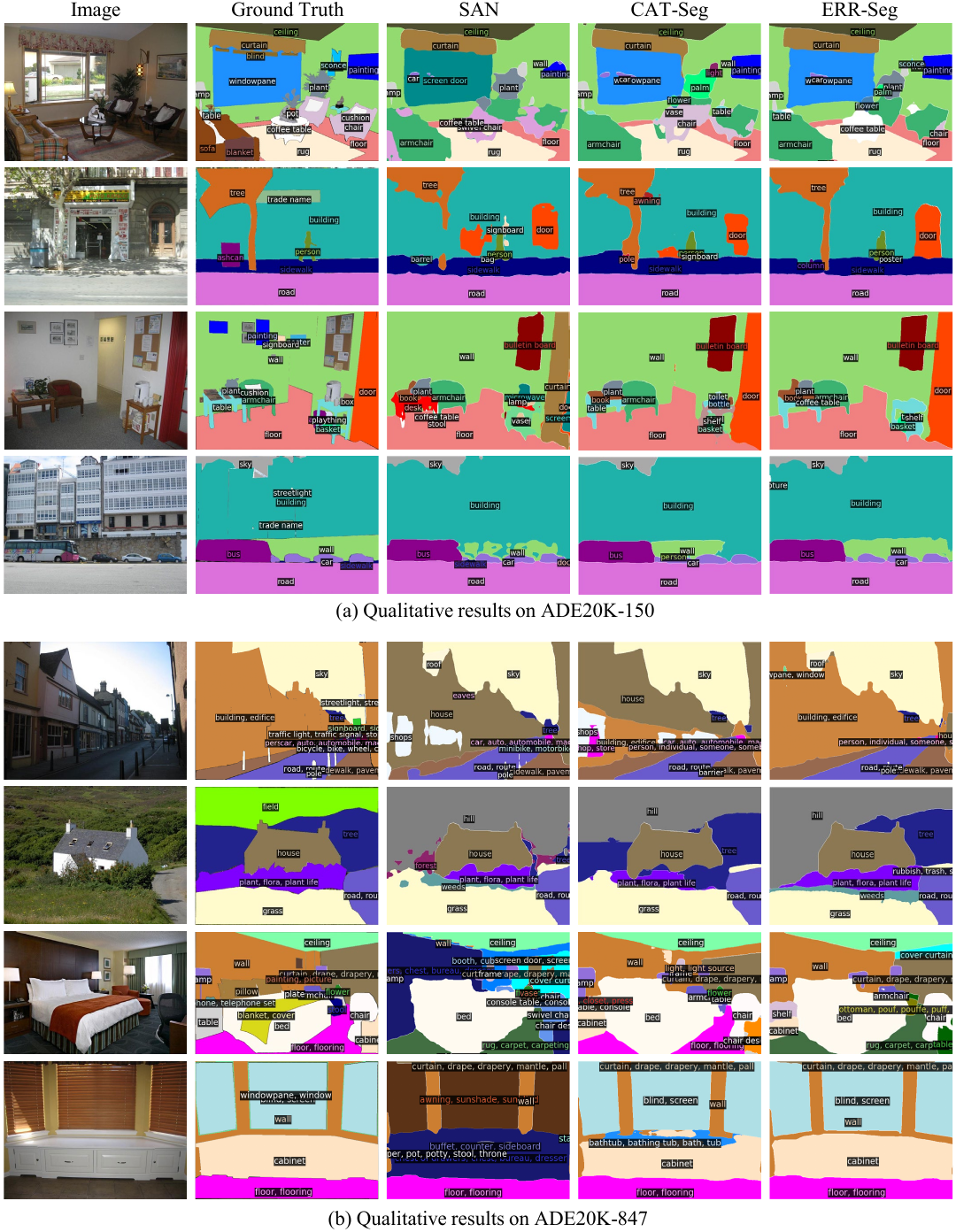}
\caption{Visualization of segmentation results on ADE20K-150 and ADE20K-847. The results of previous state-of-the-art methods SAN~\cite{xie2024sed} and CAT-Seg~\cite{cho2024cat} are also included for comparison.}
\label{fig:quality}
\end{figure*}

\subsubsection{Implementation details}
We adopt the Convolutional CLIP~\cite{cherti2023reproducible} from OpenCLIP~\cite{ilharco_gabriel_2021_5143773}, where the visual encoder is ConvNeXt~\cite{liu2022convnet} architecture. Our code is implemented on the top of Detectron2~\cite{wu2019detectron2}. The AdamW~\cite{loshchilov2017decoupled} optimizer is employed, with an initial learning rate of $2\times10^{-4}$ and a weight decay of $1\times10^{-4}$. The learning rate is set to $2\times10^{-6}$ for fine-tuning the CLIP model. {We set $\lambda=0.1,k=3$ for RRHC and $r_1=2, r_2=2$ for RRCA.} Our model is trained on COCO-Stuff using cross-entropy loss for $80$K iterations, with a batch size of $4$ and a crop size of $640\times640$. $4$ NVIDIA RTX 3090 GPUs were used for training.

\begin{table*}
\centering
\small
\caption{Comparison with state-of-the-art methods. Img. and Txt. indicate whether the VLM's image and text encoders are fine-tuned, respectively. A.B. signifies whether an additional backbone is employed.}
\resizebox{2\columnwidth}{!}{
\begin{tabular}{l|cccc|c|ccccc}
\toprule
  \multicolumn{1}{l|}{Method} &
  \multicolumn{1}{c}{VLM} &
  \multicolumn{1}{c}{Img.} &
  \multicolumn{1}{c}{Txt.} &
  \multicolumn{1}{c|}{A.B.} &
  \multicolumn{1}{c|}{Training dataset} &
  \multicolumn{1}{c}{A-847} &
  \multicolumn{1}{c}{PC-459} &
  \multicolumn{1}{c}{A-150} &
  \multicolumn{1}{c}{PC-59} &
  \multicolumn{1}{c}{PAS-20} \\ \midrule

ZegFormer~\cite{ding2022decoupling}   & ViT-B/16   & \ding{55} & \ding{55} & \checkmark     & COCO-Stuff     & 5.6  & 10.4 & 18.0 & 45.5 & 89.5 \\
SimBaseline~\cite{xu2022simple} & ViT-B/16   & \ding{55} & \ding{55} & \checkmark      & COCO-Stuff     & 7.0  & -    & 20.5 & 47.7 & 88.4 \\
OVSeg~\cite{liang2023open}       & ViT-B/16   & \checkmark & \ding{55} & \checkmark     & COCO-Stuff     & 7.1  & 11.0 & 24.8 & 53.3 & 92.6 \\
DeOP~\cite{han2023open}      & ViT-B/16   & \ding{55}  & \ding{55} & \checkmark     & COCO-Stuff-156 & 7.1  & 9.4  & 22.9 & 48.8 & 91.7 \\
SAN~\cite{xu2023side}        & ViT-B/16   & \ding{55} & \ding{55} & \ding{55}               & COCO-Stuff     & 10.1 & 12.6 & 27.5 & 53.8 & 94.0 \\
EBSeg~\cite{shan2024open}    & ViT-B/16   & \ding{55} & \ding{55}  & \checkmark              & COCO-Stuff     & 11.1 & 17.3 & 30.0 & 56.7 & 94.6 \\
SED~\cite{xie2024sed}       & ConvNeXt-B & \checkmark & \ding{55} & \ding{55}              & COCO-Stuff     & 11.4 & 18.6 & 31.6 & 57.3 & 94.4 \\
{SCAN~\cite{liu2024scan}} & {ViT-B/16} & {\ding{55}} & {\ding{55}} & {\checkmark} & {COCO-Stuff} & {\underline{13.2}} & {10.8} & {30.8} & {\underline{58.4}} & {\textbf{97.0}} \\
{MROVSeg~\cite{zhu2024mrovseg}}    & {ViT-B/16}   & {\ding{55}} & {\ding{55}}   & {\ding{55}}             & {COCO-Stuff}     & {12.1} & {19.6} & {32.0} & {\textbf{58.5}} & {95.5} \\
CAT-Seg~\cite{cho2024cat}     & ViT-B/16   & \checkmark & \checkmark  & \ding{55}              & COCO-Stuff     & 12.0 & \underline{19.0} & 31.8 & 57.5 & 94.6 \\
ERR-Seg (ours)       & ConvNeXt-B & \checkmark & \ding{55}  & \ding{55}              & COCO-Stuff     & 11.7 & 18.9 & \underline{32.9} & 56.6 & 94.9 \\
ERR-Seg (ours)        & ConvNeXt-B & \checkmark & \checkmark  & \ding{55}              & COCO-Stuff     & \textbf{13.9} & \textbf{20.5} & \textbf{35.8} & 56.9 & \underline{95.4} \\ \midrule
OVSeg~\cite{liang2023open}       & ViT-L/14   & \checkmark & \ding{55} & \checkmark          & COCO-Stuff     & 9.0  & 12.4 & 29.6 & 55.7 & 94.5 \\
ODISE~\cite{xu2023open}      & ViT-L/14   & \ding{55} & \ding{55} & \checkmark & COCO Panoptic  & 11.1 & 14.5 & 29.9 & 57.3 & - \\
SAN~\cite{xu2023side}        & ViT-L/14   & \ding{55} & \ding{55}   & \ding{55}             & COCO-Stuff     & 12.4 & 15.7 & 32.1 & 57.7 & 94.6 \\
FC-CLIP~\cite{yu2024convolutions}    & ConvNeXt-L & \ding{55} & \ding{55}  & \ding{55}              & COCO Panoptic     & 14.8 & 18.2 & 34.1 & 58.4 & 95.4 \\
Unpair-Seg~\cite{wang2024open}      & ConvNext-L  & \ding{55} & \ding{55} & \ding{55} & Merged-130K       & 14.6  & 19.5   & 32.6  & 52.2  & 93.0   \\
% SAM-CP~\cite{chen2024sam}           & ConvNext-L & \ding{55} & \ding{55} & \checkmark & COCO-Panoptic     & -     & -      & 31.8  & -     & -      \\
EBSeg~\cite{shan2024open}    & ViT-L/14   & \ding{55} & \ding{55} & \checkmark                & COCO-Stuff     & 13.7 & 21.0 & 32.8 & 60.2 & 96.4 \\
SED~\cite{xie2024sed}      & ConvNeXt-L & \checkmark & \ding{55}    & \ding{55}            & COCO-Stuff     & 13.9 & 22.6 & 35.2 & 60.6 & 96.1 \\
{MAFT+~\cite{jiao2024maftp}}    & {ConvNeXt-L}   & {\checkmark} & {\ding{55}}   & {\ding{55}}             & {COCO-Stuff}     & {15.1} & {21.6} & {36.1} & {59.4} & {96.5} \\
{OVSNet~\cite{liu2025stepping}} & {ConvNeXt-L} & {\ding{55}} & {\ding{55}} & {\ding{55}} & {COCO-Panoptic} & {\underline{16.2}} & {23.5} & {37.1} & {\underline{62.0}} & {\underline{96.9}} \\
CAT-Seg~\cite{cho2024cat}    & ViT-L/14   & \checkmark & \checkmark   & \ding{55}             & COCO-Stuff     & 16.0 & \underline{23.8} & \underline{37.9} & \textbf{63.3} & \textbf{97.0} \\
ERR-Seg (ours)        & ConvNeXt-L & \checkmark & \ding{55}  & \ding{55}              & COCO-Stuff     & 15.8 & 22.7 & 36.5 & 60.7 & 96.6 \\ 
ERR-Seg (ours)        & ConvNeXt-L & \checkmark & \checkmark  & \ding{55}              & COCO-Stuff     & \textbf{16.9} & \textbf{23.9} & \textbf{38.3} & 60.7 & 96.6 \\ 
\bottomrule
\end{tabular}
}
\label{tab:sota-compare}
\end{table*}

\begin{table}
\centering
\setlength{\tabcolsep}{1mm}
\small
\caption{Efficiency comparison. The latency (ms) is computed on a single NVIDIA RTX 3090 GPU.}
\resizebox{\columnwidth}{!}{
\begin{tabular}{lcccccccc}
\toprule
\multirow{2}{*}{Method} & \multicolumn{3}{c}{A-847} & \multicolumn{3}{c}{A-150} \\ \cmidrule(lr){2-4} \cmidrule(lr){5-7} 
            & Latency &GFLOPs & mIoU & Latency &GFLOPs & mIoU \\ \midrule
ZegFormer   & 158  & 1954  & 4.9  & 154  & 1932  & 16.9   \\
SimBaseline & 331  & 1878  & 7.0  & 329  & 1783  & 20.5   \\
DeOP        & 503  & 2791  & 7.1  & 178  & 711  & 22.9   \\
OVSeg       & 331  & 1791  & 7.1  & 324  & 1783  & 24.8   \\
SAN         & \textbf{34}   & \textbf{120} & \textbf{10.1}   & \textbf{32} & \textbf{95}   & \textbf{27.5} \\ \midrule
SED         & 196  & 571 & 11.4 & 114  & 291  & 31.6  \\
CAT-Seg     & 552  & 2154  & 12.0 & 369  & 1516  & 31.8  \\
ERR-Seg        & \textbf{114}  & \textbf{348} & \textbf{13.9}   & \textbf{85} & \textbf{288}   & \textbf{35.8} \\ \bottomrule
\end{tabular}
}
\label{tab:efficient-compare}
\end{table}

\subsection{Main Results}
\subsubsection{Comparing with state-of-the-art}
We compare our proposed ERR-Seg with existing state-of-the-art methods on the five standard evaluation settings, with the results presented in Table~\ref{tab:sota-compare}. To ensure a fair comparison, all models utilize CLIP as the VLM. We also include whether CLIP’s image and text encoders are fine-tuned. When both the image and text encoders are fine-tuned, ERR-Seg achieves remarkable performance on these five settings, particularly in the base model. {Notably, on A-847, PC-459, and A-150, ERR-Seg achieves significant gains, surpassing the previous best results by margins of $5.3\%$, $7.9\%$, and $11.9\%$, respectively. In the large model setting, ERR-Seg continues to outperform prior arts, exceeding the best previous scores on A-847, PC-459, and A-150 by $4.3\%$, $0.4\%$, and $1.1\%$, respectively.} When only the text encoder is fine-tuned, ERR-Seg also outperforms SED~\cite{xie2024sed} on all five settings, with an improvement of $13.7\%$, $0.4\%$, $5.2\%$, $0.2\%$ and $0.6\%$ on A-847, PC-459, A-150, PC-59 and PAS-20, respectively.

{However, ERR-Seg does not perform as well as CAT-Seg on PC-59, exhibiting a $4.1\%$ lower performance than CAT-Seg in the large model. As discussed in Section~\ref{analysis_data}, PC-59 primarily assesses in-domain open-vocabulary performance. This indicates that CAT-Seg might fit more of the seen categories during training, thus showcasing superior in-domain performance compared to our model but lagging in out-domain performance. In contrast, with the introduction of redundant class reduction, ERR-Seg customizes the vocabulary list for each image during training, which helps prevent overfitting to the seen categories. A similar trend is observed with SCAN: it outperforms CAT-Seg by 0.9 and 2.4 on PC-59 and PAS-20, respectively, but suffers a significant performance drop of 8.2 on PC-459.
}

\begin{table}[htbp]

\caption{Comparison on seen and unseen classes.}
\label{tab:unseen}
\resizebox{\columnwidth}{!}{
\begin{tabular}{lcccccc}
\toprule
\multirow{2}{*}{Method} & \multicolumn{2}{c}{A-847} & \multicolumn{2}{c}{PC-459} & \multicolumn{2}{c}{A-150} \\ \cmidrule(lr){2-3} \cmidrule(lr){4-5} \cmidrule(lr){6-7} 
                        & Seen       & Unseen       & Seen        & Unseen       & Seen       & Unseen       \\ \midrule
CAT-Seg                 & 29.2       & 9.6          & 38.3        & 12.5         & 42.1       & 24.4         \\
ERR-Seg                 & \textbf{30.1}       & \textbf{11.6}         & \textbf{40.1}        & \textbf{13.9}         & \textbf{44.7}       & \textbf{29.4}         \\ \bottomrule
\end{tabular}}
\end{table}

{
Tab.~\ref{tab:unseen} presents a comparison with CAT-Seg on both seen and unseen classes. Here, unseen classes are defined as those with a label similarity exceeding 0.95 to the COCO-Stuff training set, based on the analysis in Sec.~\ref{analysis_data}. The results demonstrate that our method not only generalizes effectively from seen to unseen classes, but also achieves consistent improvements across both settings. Furthermore, our method exhibits more substantial gains on unseen classes relative to CAT-Seg: on the seen classes of A-847, PC-459, and A-150, the relative improvements are $3.1\%$, $4.7\%$, and $6.2\%$, respectively, while on unseen classes, the relative improvements reach $20.8\%$, $11.2\%$, and $20.5\%$.
}

\subsubsection{Efficiency comparison}
Table~\ref{tab:efficient-compare} presents the results of the latency comparisons. Although cost-based CAT-Seg shows notable performance enhancement over mask-based SAN, it also results in a considerable increase in latency. In contrast, our proposed ERR-Seg not only further improves the performance of cost-based methods but also reduces their latency significantly. Specifically, the latency is reduced by $79.3\%$ on A-847 and $77.0\%$ on A-150.

\subsubsection{Qualitative results}
Fig.~\ref{fig:quality} presents a qualitative comparison among SAN~\cite{xu2023side}, CAT-Seg~\cite{cho2024cat}, and our proposed ERR-Seg on the validation sets of ADE20K-150 and ADE20K-847. The results highlight the superiority of our method over previous approaches in mask generation and open-vocabulary recognition. In addition, ERR-Seg can identify targets that were previously overlooked in ADE20K. While the current OVSS model shows remarkable open-vocabulary recognition capability due to the advancements of VLMs, it struggles with segmenting intricate masks. There is still potential for improvement in effectively recognizing small and nested targets.

\subsection{Ablation Study and Analysis}

\subsubsection{Ablation study on RRHC}
Table~\ref{tab:rrhc} presents the ablation study on RRHC. The results indicate that removing the class redundancy leads to performance increases on some benchmarks, such as $+0.4$ mIoU on PC-459 and $+0.5$ mIoU on A-150. This demonstrates that our proposed class redundancy removal strategy can serve not only as a methodology to alleviate computation overload for OVSS but also to improve performance. Moreover, it also demonstrates the effectiveness of our hierarchical cost maps, which leads to $+1.2$ mIoU improvement on PC-459.

Without RRHC, the gradient for updating stage 4 of CLIP's visual encoder is derived solely from $\mathcal{F}_4^s$. However, when using RRHC, the gradient can be sourced from $\mathcal{F}_3^s$ and $\mathcal{F}_4^s$, as it functions as a parallel branch. Table~\ref{tab:rrhc} investigates the impact of different gradient sources on updating stage 4. The results show that relying solely on $\mathcal{F}_3^s$ leads to notably poorer model performance. Furthermore, the combination of gradients from $\mathcal{F}_3^s$ and $\mathcal{F}_4^s$ yields slightly better performance on some benchmarks than relying solely on $\mathcal{F}_4^s$.

\begin{table}[!H]
\centering
\setlength{\tabcolsep}{1mm}
\small
\caption{Ablation study and on RRHC.}
\label{tab:rrhc}
\resizebox{\columnwidth}{!}{
\begin{tabular}{c|ccccc}
\toprule
Class redundancy                  & A-847 & PC-459 & A-150 & PC-59 & PAS-20 \\ \midrule
$w/o$ remove                      & \textbf{14.0} & 20.1   & 35.3  & 56.8   & 94.9   \\
$w/$  remove                      & 13.9  & \textbf{20.5} & \textbf{35.8} & \textbf{56.9} & \textbf{95.4}   \\ \midrule
Cost maps                         & A-847 & PC-459 & A-150 & PC-59 & PAS-20 \\ \midrule
Single                            & 13.3  & 19.3   & 35.5  & 56.4  & 95.0   \\
Hierarchical                      & \textbf{13.9}  & \textbf{20.5} & \textbf{35.8} & \textbf{56.9} & \textbf{95.4}   \\ \midrule
Stage 4 gradient from             & A-847 & PC-459 & A-150 & PC-59 & PAS-20 \\ \midrule
$\mathcal{F}_3^s$                 & 7.3   & 13.5   & 20.1  & 45.5  & 90.6   \\
$\mathcal{F}_4^s$                 & \textbf{14.1}  & 20.4   & 35.6  & \textbf{57.0}  & 94.7   \\
$\mathcal{F}_3^s+\mathcal{F}_4^s$ & 13.9  & \textbf{20.5}   & \textbf{35.8}  & 56.9  & \textbf{95.4} \\ \bottomrule
\end{tabular}
}
\vspace{-10pt}
\end{table}

\subsubsection{Hyperparameter analysis for RRHC}
\begin{figure}[htbp]
\centering
\includegraphics[width=\columnwidth]{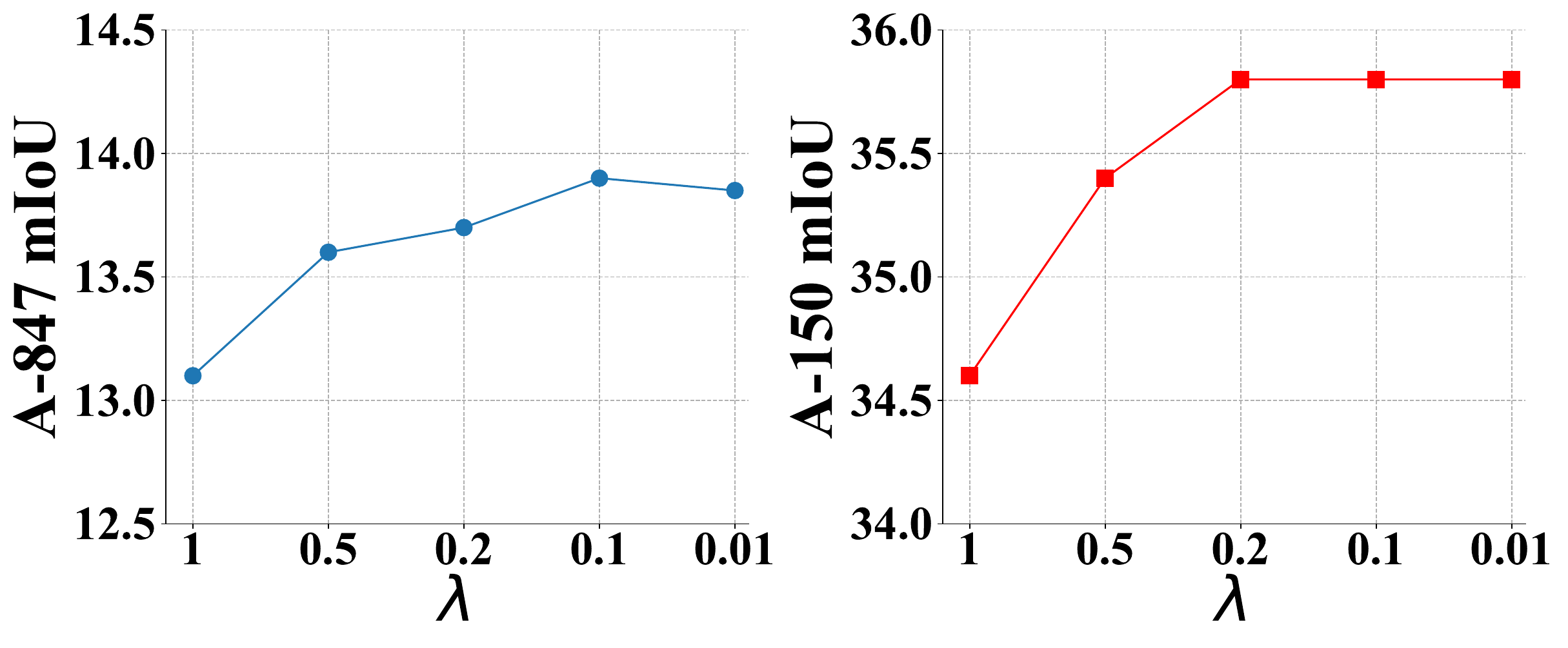}
\vspace{-20pt}
\caption{Impact of score coefficient $\lambda$.}
\label{fig:gamma}
\vspace{-10pt}
\end{figure}

Fig.~\ref{fig:gamma} illustrates how the score coefficient $\lambda$ influences class scoring, with optimal performance achieved at $\lambda=0.1$. A larger $\lambda$ may blur the distinction between classes with similar rankings, while a smaller $\lambda$ might overlook certain classes.

Table~\ref{tab:RCR} presents the effect of the top $k$ categories of a pixel, indicating that the redundant class reduction mechanism is minimally sensitive to $k$, with slightly superior results observed at $k=3$. A smaller $k$, such as 1, risks overlooking existing categories due to potential inaccuracies in cost maps' prior knowledge. Conversely, a larger $k$ may introduce redundant categories, hindering its ability to identify the most relevant classes.

\begin{figure}[htbp]
\centering
\includegraphics[width=0.67\columnwidth]{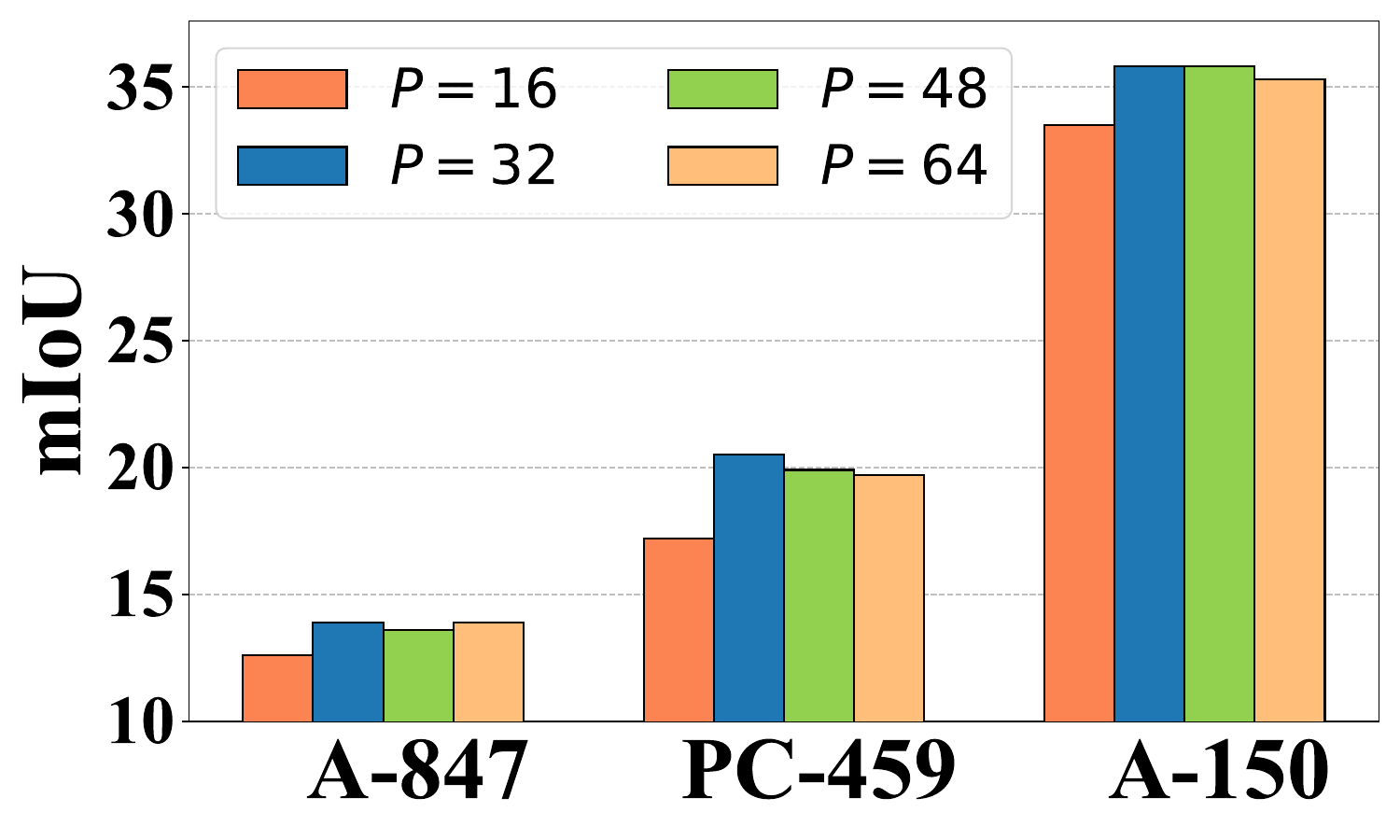}
\caption{Analysis of selected class number $P$ during training. It indicates that excessively large or small values of $P$ result in performance degradation.}
\label{fig:p}
\end{figure}

Fig.~\ref{fig:p} explores the effect of the selected class number $P$ during training on the COCO-Stuff dataset, highlighting that setting $P=32$ yields optimal performance. It also reveals that setting $P$ either too large or too small can hinder the modeling of contextual information.  During inference, $P$ is set to the original number of classes of each dataset to ensure a fair comparison.

Table~\ref{tab:K} illustrates the influence of the sampled class number $P$ in the redundant class reduction mechanism during the inference phase. It is evident that as $P$ increases gradually, the performance tends to reach a saturation point. The performance is observed to saturate as $P$ increases, consistent with the expectation that a sufficiently large $P$ can encompass all relevant categories. Specifically, we assign the following values to $P$: 48 for A-847, 48 for PC-459, 32 for A-150, 24 for PC-59, and 16 for PAS-20.

\begin{table}[h]
\centering
\setlength{\tabcolsep}{1mm}
\small
\caption{Effect of Top $k$ categories.}
\label{tab:RCR}
\begin{tabular}{c|ccccc}
\toprule
Top $k$ categories       & A-847 & PC-459 & A-150 & PC-59 & PAS-20 \\ \midrule
1                        & 13.8  & 20.5   & 35.8  & 56.9  & 95.4   \\
3                        & 13.9  & 20.5   & 35.8  &56.9   & 95.4   \\
5                        & 13.8  & 20.5   & 35.8  & 56.8  & 95.4   \\ \bottomrule
\end{tabular}
\vspace{-10pt}
\end{table}

\begin{table}[htbp]
\centering
\setlength{\tabcolsep}{1mm}
\small
\caption{The impact of sampled class number $P$ during inference.}
\resizebox{\columnwidth}{!}{
\begin{tabular}{cccccccccc}
\toprule
\multicolumn{2}{c}{A-847} & \multicolumn{2}{c}{PC-459} & \multicolumn{2}{c}{A-150} & \multicolumn{2}{c}{PC-59} & \multicolumn{2}{c}{PAS-20} \\ \cmidrule(lr){1-2} \cmidrule(lr){3-4} \cmidrule(lr){5-6}  \cmidrule(lr){7-8}  \cmidrule(lr){9-10}
P$\downarrow$  & mIoU$\uparrow$ & P$\downarrow$  & mIoU$\uparrow$ & P$\downarrow$  & mIoU$\uparrow$ & P$\downarrow$  & mIoU$\uparrow$ & P$\downarrow$  & mIoU$\uparrow$ \\ \midrule
32 & 13.6 & 16 & 18.8 & 16 & 34.9 & 8  & 54.9 & 4  & 90.2 \\
48 & 13.9 & 32 & 19.9 & 24 & 35.5 & 16 & 56.9 & 8  & 95.3 \\
64 & 13.9 & 48 & 20.5 & 32 & 35.8 & 24 & 56.9 & 12 & 95.4 \\
80 & 13.9 & 64 & 20.5 & 40 & 35.8 & 32 & 57.0 & 16 & 95.4 \\ \bottomrule
\end{tabular}
}
\label{tab:K}
\end{table}

{
\subsubsection{Analysis on RRHC}
Table~\ref{tab:rrhc_analysis} presents the impact of removing redundant classes in RRHC on the final segmentation performance. We report the average number of classes per image in the ground truth, as well as the number of predicted classes overlapping with the ground truth, both before and after redundant class removal. It can be observed that the removal leads to a slight reduction in the number of recognized classes, for instance, on A-150, an average of 0.2 classes per image is no longer detected. Nevertheless, this minor decline is a worthwhile trade-off, as removing redundant classes brings gains in both segmentation accuracy (e.g., 0.4 and 0.5 improvement on PC-459 and A-150, respectively) and inference speed (discussed in Sec.~\ref {sec:analysis}).
}

\begin{table}[htbp]
\centering
\setlength{\tabcolsep}{1mm}
\caption{Analysis of redundant class removal in RRHC. The ``\#Classes'' columns denote the average number of predicted classes that overlap with the ground truth per image.}
\label{tab:rrhc_analysis}
\resizebox{\columnwidth}{!}{
\begin{tabular}{lcccccc}
\toprule
\multirow{2}{*}{Method} & \multicolumn{2}{c}{A-847} & \multicolumn{2}{c}{PC-459} & \multicolumn{2}{c}{A-150} \\ \cmidrule(lr){2-3} \cmidrule(lr){4-5} \cmidrule(lr){6-7} 
                        & \#Classes        & mIoU       & \#Classes        & mIoU        & \#Classes        & mIoU       \\ \midrule
Ground Truth            & 10.2         & -          & 6.5          & -           & 8.5          & -          \\
$w/o$ removal              & 6.2          & 14.0       & 4.7          & 20.1        & 6.1          & 35.3       \\
$w/$  removal              & 5.7          & 13.9       & 4.4          & 20.5        & 5.9          & 35.8       \\ \bottomrule
\end{tabular}
}
\end{table}

\subsubsection{Ablation study on RRCA}
RRCA aims to incorporate more contextual information at both the class-level and spatial-level within the cost maps. We first conduct experiments to analyze the spatial-level and class-level aggregation separately. We observed a significant decrease in performance when the spatial-level aggregation and the class-level aggregation are removed, as shown in Table~\ref{tab:escf}. {Table~\ref{tab:ss} presents the effects of the sequence reduction strategies, SSR and CSR, including their impact on GFLOPs, memory usage (GiB) and latency (ms).} The findings indicate that the use of CSR or SSR alone does not negatively affect performance. {Overall, equipping SSR and CSR achieves a reduction of $49.6\%$ GFLOPs, $55.6\%$ memory and $52.4\%$ latency without compromising performance.} Additionally, Table~\ref{tab:escf-star} compares the effect of different MLP architectures within the cost aggregation blocks. Here, Vanilla refers to the standard MLP consisting of two linear layers and an activation function, while the Modified MLP incorporates the star operation to strengthen its representational capacity. It shows that the performance of ERR-Seg is improved on four datasets by adopting the modified MLP.

\begin{table}[htbp]
\centering
\caption{The impact of spatial-level aggregation and class-level aggregation.}
\resizebox{\columnwidth}{!}{
\begin{tabular}{cc|ccccc}
\toprule
Spatial       & Class & A-847 & PC-459 & A-150 & PC-59 & PAS-20 \\ \midrule
              &       & 9.3  & 12.8          & 26.1          & 43.6          & 90.8          \\
\checkmark    &       &13.1           & 19.2            & 34.4           & 56.3           & 94.4            \\
&  \checkmark         &13.0           & 19.6            & 34.1           & 55.4           & \textbf{95.5}   \\
\checkmark    & \checkmark    &\textbf{13.9}  & \textbf{20.5}   & \textbf{35.8}  & \textbf{56.9}  & 95.4     \\ \bottomrule
\end{tabular}
}
\label{tab:escf}
\end{table}

\begin{table}[h]
\centering

\setlength{\tabcolsep}{1mm}
\small
\caption{Effect of the Spatial-level Sequence Reduction (SSR) and Class-level Sequence Reduction (CSR).}
\resizebox{\columnwidth}{!}{
\begin{tabular}{cc|cccc|ccc}
\toprule
SSR & CSR & A-847 & PC-459 & A-150 & PC-59  & GFLOPs & Memory & Latency\\ \midrule
   &        &  13.5    & 20.0   & 35.4  & 57.0      & 188.1  & 9.5  & 103       \\
           & \checkmark        & 14.2  & 20.0   & 35.9  & 56.9   & 148.9  & 7.8  & 91       \\
\checkmark   &       & 14.2     & 19.9   & 35.6   & 57.0  & 134.6  & 5.9  & 57       \\
\checkmark          & \checkmark        & 13.9  & 20.5   & 35.8  & 56.9   & \textbf{95.3}  & \textbf{4.2}  &  49      \\ \bottomrule
\end{tabular}
}
\label{tab:ss}
\end{table}

\begin{table}[h]
\centering
\small
\caption{Impact of different MLP structures in cost aggregation blocks.}
\resizebox{\columnwidth}{!}{
\begin{tabular}{l|ccccc}
\toprule
MLP          & A-847 & PC-459 & A-150 & PC-59 & PAS-20 \\ \midrule
Vanilla      & \textbf{14.1}          & 20.1          & 35.6          & 56.3          & 94.7          \\
Modified     & 13.9 & \textbf{20.5} & \textbf{35.8} & \textbf{56.9} & \textbf{95.4} \\ \bottomrule
\end{tabular}
}
\label{tab:escf-star}
\end{table}

\subsubsection{Impact of CLIP architecture}
Following previous methods~\cite{xie2024sed, yu2024convolutions}, our ERR-Seg also adopts the hierarchical CLIP (ConvNeXt) instead of the plain CLIP (ViT). The impact of the CLIP architecture is illustrated in Table~\ref{tab:clip-arch}. To ensure a fair comparison, training and evaluation are conducted using an input size of $384^2$ with frozen CLIP. The results show that employing hierarchical CLIP leads to a +$2.2$ +and +$4.1$ mIoU improvement on A-847 and A-150, respectively. {Furthermore, compared to the ViT-based architecture, it slightly reduces inference latency, with an 8 ms reduction on ADE20K-150.} In comparison to plain CLIP, hierarchical CLIP offers the following advantages: (1) it allows the use of middle-layer features to enhance the reconstruction of high-resolution semantic maps, and (2) it permits direct inference on high-resolution images, eliminating the need for the sliding window inference strategy~\cite{cho2024cat} employed in plain CLIP.

\begin{table}[htbp]
\centering

\setlength{\tabcolsep}{1mm}
\caption{Impact of CLIP Architecture.}
\resizebox{\columnwidth}{!}{
\begin{tabular}{l|c|ccccc}
\toprule
CLIP Arch. & Latentcy & A-847 & PC-459 & A-150 & PC-59 & PAS-20 \\ \midrule
ViT-B/16   & 56       & 7.5   & 13.0   & 22.8  & 49.7  & 91.0   \\
ConvNeXt-B & 47       & 9.7   & 12.5   & 26.9  & 49.4  & 89.9   \\ \bottomrule
\end{tabular}
}
\label{tab:clip-arch}
\end{table}

\subsubsection{Impact of input size}
Table~\ref{tab:size} presents the results with varying input sizes during the inference phase. The findings highlight a decrease in performance when input sizes are too large or too small. Specifically, for models based on ConvNeXt-B and ConvNeXt-L, the recommended input sizes are $640^2$ and $768^2$, respectively. Our analysis reveals that using ConvNeXt-L as the backbone of CLIP, there is a better alignment between the image and text spaces, enabling the extraction of more refined pixel-text cost maps and thus supporting larger input sizes.

\begin{table}[t!]
\centering
\setlength{\tabcolsep}{1mm}
\small
\caption{The impact of crop size during inference.}
\resizebox{\columnwidth}{!}{
\begin{tabular}{l|l|ccccc}
\toprule
Backbone                    & Size     & A-847         & PC-459        & A-150         & PC-59         & PAS-20        \\ \midrule
\multirow{4}{*}{ConvNeXt-B} & $512^2$  & \underline{13.7}    & 19.8          & 35.1          & 56.7          & \underline{95.3}    \\
                            & \textbf{$640^2$}  & \textbf{13.9} & \textbf{20.5} & \textbf{35.8} & \underline{56.9}    & \textbf{95.4} \\
                            & $768^2$  & 13.5          & \underline{20.1}    & \underline{35.7}    & \textbf{57.1} & 95.1          \\
                            & $1024^2$ & 13.1          & 18.7          & 34.3          & 55.5          & 93.3          \\\midrule \midrule
\multirow{4}{*}{ConvNeXt-L} & $512^2$  & 16.0          & 22.9          & 36.9          & 60.0          & 96.7          \\ 
                            & $640^2$  & \underline{16.6}    & \underline{23.5}    & 37.8          & \underline{60.5} & \textbf{96.8}    \\
                            & $768^2$  & \textbf{16.9} & \textbf{23.9} & \textbf{38.3} & \textbf{60.7} & \underline{96.6} \\
                            & $1024^2$ & 16.6          & 23.4          & \underline{38.2}    & 59.8          & 96.2        \\ \bottomrule
\end{tabular}
}
\label{tab:size}
\end{table}

{
\subsubsection{Failure case analysis.}
Fig.~\ref{fig:quality} presents a qualitative analysis of ERR-Seg's performance. The results indicate that our model can identify small objects, such as the person and cars depicted in Fig.~\ref{fig:quality} (a). However, the model struggles to segment precise masks for extremely narrow or small objects, such as the table legs and road railings shown in Fig.~\ref{fig:quality} (a) and (b). Moreover, ERR-Seg demonstrates considerable robustness in parsing complex scenes, such as cluttered indoor environments. Nevertheless, it remains susceptible to semantic misclassification, for instance, confusing a sofa with an armchair in Fig.~\ref{fig:quality} (a). It is worth noting that addressing these issues remains a key objective in the field of semantic segmentation.
}

\section{Conclusion}
In conclusion, we introduced ERR-Seg, an efficient framework for open-vocabulary semantic segmentation. We identified that the redundancy in cost-based methods not only increases computational latency but also hinders the attention mechanism from capturing spatial-level and class-level dependencies. To address this, ERR-Seg incorporates the Redundancy-Reduced Hierarchical Cost maps (RRHC), which eliminates irrelevant classes using semantic priors from VLMs and incorporates hierarchical cost maps for enriched semantic representation. Furthermore, the Redundancy-Reduced Cost Aggregation (RRCA) significantly reduces computational overhead via sequence compression. Experimental results demonstrate that ERR-Seg establishes a new state-of-the-art in terms of efficiency and accuracy.

{
\noindent \textbf{Limitations and future work.}
Our failure case analysis reveals two primary limitations of the proposed OVSS framework: difficulties in generating precise masks for extremely small or narrow objects and inaccuracies in class recognition. We attribute these challenges to the cost aggregation paradigm's reliance on CLIP for generating dense pixel-text cost maps, which provides insufficient fine-grained visual representations for these challenging scenarios. Future work will enhance the OVSS framework by integrating powerful vision foundation models (e.g., DINO~\cite{simeoni2025dinov3}, SAM~\cite{kirillov2023segment}) to improve mask quality and leveraging advanced Multimodal Large Language Models (e.g., Qwen-VL series~\cite{Qwen-VL, Qwen2-VL, Qwen2.5-VL}) to provide more robust open-vocabulary recognition capability. Moreover, future work also will focus on designing lightweight learnable class selection module to automatically adapt to diverse vocabularies and image inputs.
}

\appendix
\section{Appendix}
\begin{figure*}[htbp]
\centering
\includegraphics[width=2\columnwidth]{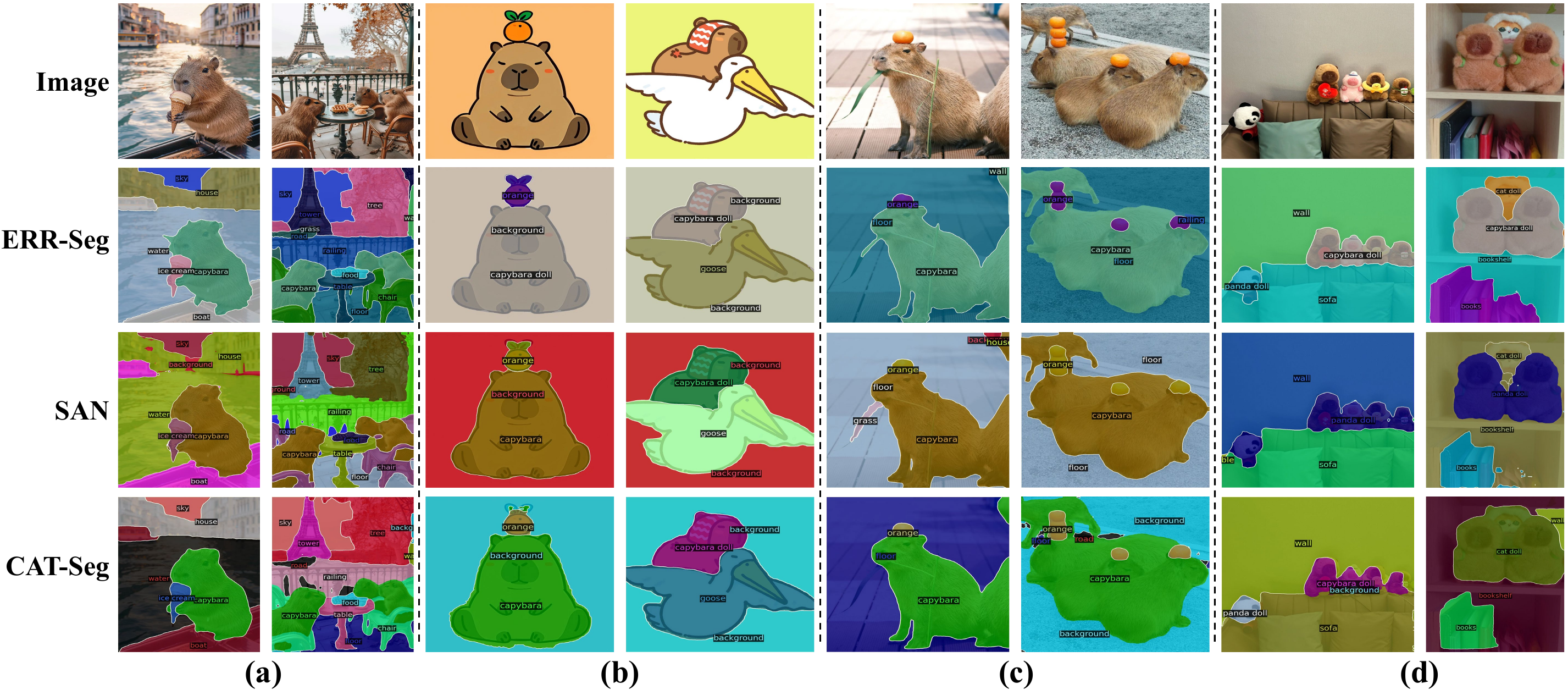}
\caption{Visualization of segmentation results in various domains. Our proposed ERR-Seg is capable of segmenting capybaras (a rare category in public datasets) from various domains, including (a) synthesized images, (b) cartoon images, (c) natural images, and (d) capybara dolls. Moreover, ERR-Seg achieves more precise masks than SAN~\cite{xu2023side} and CAT-Seg~\cite{cho2024cat}.}
\label{fig:all_capy}
\end{figure*}

\subsection{Proof for Proposition 1}
\begin{proof}
By utilizing Eq.~\ref{er} and Eq.~\ref{q}, the expression $\Delta r=O^{r}(q)/O^{e,r}(q)$ can be simplified as:
\begin{align}
\scalebox{0.85}{$\displaystyle
\Delta _r=\sum_{k=M+1}^K{\exp \left( \frac{A_{qk}}{\sqrt{C}} \right) S(k)}/\sum_{k=1}^K{\exp \left( \frac{A_{qk}}{\sqrt{C}} \right) S(k)}.
$}
\end{align}
Similarly, $\Delta _p$ can be simplified as:
\begin{align}
\scalebox{0.85}{$\displaystyle
\Delta _p=\sum_{k=M+1}^P{\exp \left( \frac{B_{qk}}{\sqrt{C}} \right) S(k)}/\sum_{k=1}^P{\exp \left( \frac{B_{qk}}{\sqrt{C}} \right) S(k)},
$}
\end{align}
where $B \in \mathbb{R}^{P\times P}$ represents the attention scores matrix after redundancy reduction. Let us define:
\begin{align}
\Delta _{r}^{\prime}=\left( \frac{1}{\Delta _r} -1 \right),\\
\Delta _{p}^{\prime}=\left( \frac{1}{\Delta _p} -1 \right).
\end{align}
Since $0<\Delta _r,\Delta _p<1$, it follows that $\Delta _{r}^{\prime},\Delta _{p}^{\prime}>0$. To compare $\Delta _r$ and $\Delta _p$, we first compare $\Delta _{r}^{\prime}$ and $\Delta _{p}^{\prime}$, by analyzing the following formula:
\begin{align}
\scalebox{0.85}{$\displaystyle
\frac{\Delta _{r}^{\prime}}{\Delta _{p}^{\prime}}=\frac{\sum_{k=1}^M{\exp \left( \frac{A_{qk}}{\sqrt{C}} \right) S(k)}}{\sum_{k=M+1}^K{\exp \left( \frac{A_{qk}}{\sqrt{C}} \right) S(k)}}\frac{\sum_{k=M+1}^P{\exp \left( \frac{B_{qk}}{\sqrt{C}} \right) S(k)}}{\sum_{k=1}^M{\exp \left( \frac{B_{qk}}{\sqrt{C}} \right) S(k)}}.
$}
\end{align}
Let us define:
\begin{align}
\scalebox{0.85}{$\displaystyle
\alpha =\sum_{k=1}^M{\exp \left( \frac{A_{qk}}{\sqrt{C}} \right) S(k)}/\sum_{k=1}^M{\exp \left( \frac{B_{qk}}{\sqrt{C}} \right) S(k)}
$}
\end{align}
and
\begin{align}
\scalebox{0.85}{$\displaystyle
\beta =\sum_{k=M+1}^P{\exp \left( \frac{B_{qk}}{\sqrt{C}} \right) S(k)}/\sum_{k=M+1}^K{\exp \left( \frac{A_{qk}}{\sqrt{C}} \right) S(k)}.
$}
\end{align}
Thus, we have
\begin{align}
\frac{\Delta _{r}^{\prime}}{\Delta _{p}^{\prime}}=\alpha \beta.
\end{align}
Since $A$ and $B$ arise from similarity computations between the same query and key, and redundant categories do not impact the similarity scores of common categories, we have
\begin{align}
\forall 1\le i,j\le P,A_{ij}=B_{ij}.
\end{align}
This implies that $\alpha = 1$ and $\beta < 1$. Hence,
\begin{align}
\frac{\Delta _{r}^{\prime}}{\Delta _{p}^{\prime}}=\alpha \beta <1.
\end{align}
It indicates that $\Delta _{r}^{\prime} < \Delta _{p}^{\prime}$. Thus, we can get $\Delta _p < \Delta _r$. In this way, we finish the proof.
\end{proof}

\begin{figure}[htbp]
\centering
\includegraphics[width=0.95\columnwidth]{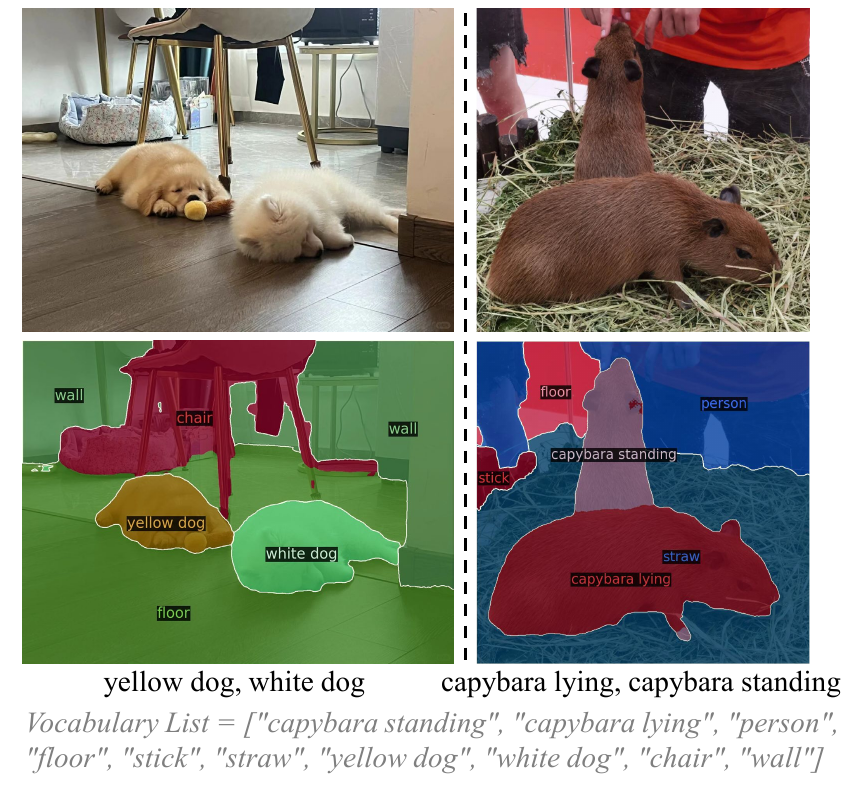}
\caption{\textbf{Visualization of ERR-Seg’s attribute understanding.} ERR-Seg can correctly distinguish between a yellow dog and a white dog and between a lying capybara and a standing capybara.}
\label{fig:cases}
\end{figure}

\subsection{More Qualitative Results}
As presented in Fig.~\ref{fig:all_capy}, ERR-Seg shows its impressive ability to segment and identify capybaras in various domains, highlighting its universal segmentation capability. The selection of capybaras is due to their rarity in public datasets. Furthermore, compared with SAN and CAT-Seg, ERR-Seg achieves more precise masks with reduced noise.

Fig.~\ref{fig:cases} presents two simple examples that illustrate the ability of ERR-Seg to understand attributes such as color and posture. Our model accurately segments dogs of various colors and capybaras in different postures, demonstrating its exceptional open-world understanding capabilities.

\newpage
\printcredits

\bibliographystyle{unsrt}
\bibliography{cas-refs}

\clearpage
%\vskip3pt

\vfill

\end{document}